\documentclass[letterpaper]{article} 
\usepackage[submission]{arxiv}  
\usepackage{times}  
\usepackage{helvet}  
\usepackage{courier}  
\usepackage[hyphens]{url}  
\usepackage{graphicx} 
\usepackage{natbib}  
\usepackage{caption} 
\usepackage{algorithm}
\usepackage{algorithmic}

\usepackage{times}
\usepackage{graphicx}
\usepackage{amsmath}
\usepackage{amssymb}
\usepackage{enumerate}
\usepackage{booktabs}
\usepackage{svg}
\usepackage{bigstrut}
\usepackage{subcaption}
\usepackage{multirow}
\usepackage{mwe}
\usepackage{diagbox}
\usepackage{hhline}
\usepackage[font=small]{caption}
\usepackage{slashbox}
\usepackage{rotating}
\usepackage{array}
\newcolumntype{P}[1]{>{\centering\arraybackslash}p{#1}}
\newcolumntype{M}[1]{>{\centering\arraybackslash}m{#1}}

\usepackage[pagebackref=false,breaklinks=true,colorlinks=true,bookmarks=false,citecolor=blue]{hyperref}
\ifdefined\nohyperref\else\ifdefined\hypersetup
  \definecolor{mydarkblue}{rgb}{0,0.08,0.45}
  \hypersetup{ %
    colorlinks=true,
    linkcolor=mydarkblue,
    citecolor=mydarkblue,
    filecolor=mydarkblue,
    urlcolor=mydarkblue,
    }
  \fi
\fi

\usepackage[capitalize]{cleveref}
\crefname{section}{Sec.}{Secs.}
\Crefname{section}{Section}{Sections}
\Crefname{table}{Table}{Tables}
\crefname{table}{Tab.}{Tabs.}

\newcommand{\vect}[1]{\mathbf{#1}}

\def\Put(#1,#2)#3{\leavevmode\makebox(0,0){\put(#1,#2){#3}}}

\newcommand{\cv}{\vect{c}}
\newcommand{\Cv}{\vect{C}}
\newcommand{\ev}{\vect{e}}

\newcommand{\fv}{\vect{f}}
\newcommand{\bv}{\vect{b}}
\newcommand{\Fv}{\vect{F}}
\newcommand{\gv}{\vect{g}}

\newcommand{\Tv}{\vect{T}}

\newcommand{\nv}{\vect{n}}
\newcommand{\sv}{\vect{s}}

\newcommand{\xv}{\vect{x}}
\newcommand{\Xv}{\vect{X}}
\newcommand{\yv}{\vect{y}}

\newcommand{\ssi}[1]{^{(#1)}}

\newcommand{\MIS}{M_{\mathsf{IS}}}
\newcommand{\MAS}{M_{\mathsf{AS}}}

\DeclareMathOperator*{\argmin}{arg\,min}

\usepackage{amsthm}

\theoremstyle{definition}

\def\ie{{i.e.}}
\def\eg{{e.g.}}

\usepackage{newfloat}
\usepackage{listings}
\DeclareCaptionStyle{ruled}{labelfont=normalfont,labelsep=colon,strut=off} 
\floatstyle{ruled}
\newfloat{listing}{tb}{lst}{}
\floatname{listing}{Listing}
\title{WrappingNet: Mesh Autoencoder via Deep Sphere Deformation}
\author{
    Eric~Lei\textsuperscript{* \dag}, Muhammad~Asad~Lodhi\textsuperscript{*}, Jiahao~Pang\textsuperscript{*}, Junghyun~Ahn\textsuperscript{*}, Dong~Tian\textsuperscript{*}
}
\affiliations{
    \textsuperscript{*}InterDigital, New York, NY, USA \\
    \textsuperscript{\dag}Dept. of Electrical and Systems Engineering, University of Pennsylvania, Philadelphia, PA, USA \\
    {\tt \small elei@seas.upenn.edu, \{muhammad.lodhi, jiahao.pang, junghyun.ahn, dong.tian\}@interdigital.com}
}

\begin{document}

\pagestyle{plain}
\maketitle

\begin{abstract}
There have been recent efforts to learn more meaningful representations via fixed length codewords from mesh data, since a mesh serves as a complete model of underlying 3D shape compared to a point cloud. However, the mesh connectivity presents new difficulties when constructing a deep learning pipeline for meshes. Previous mesh unsupervised learning approaches typically assume category-specific templates, e.g., human face/body templates. It restricts the learned latent codes to only be meaningful for objects in a specific category, so the learned latent spaces are unable to be used across different types of objects. In this work, we present WrappingNet, the first mesh autoencoder enabling general mesh unsupervised learning over heterogeneous objects. It introduces a novel \emph{base graph} in the bottleneck dedicated to representing mesh connectivity, which is shown to facilitate learning a shared latent space representing object shape. The superiority of WrappingNet mesh learning is further demonstrated via improved reconstruction quality and competitive classification compared to point cloud learning, as well as latent interpolation between meshes of different categories.
\end{abstract}

\section{Introduction}
\label{sec:intro}
Convolutional autoencoders applied on Euclidean grid-based data have shown good performance in terms of producing high-quality reconstruction, learning useful latent representations that generalize well, as well as computational efficiency. Recent works have attempted to extend these architectures to non-Euclidean domains \cite{gdl}. In this paper, we work with triangle mesh data. However, the variable size, lack of a fixed grid structure, and ample connectivity of mesh data present challenges that prevent similar autoencoder capabilities to be achieved.

\begin{figure}
    
\end{figure}

\begin{figure}
     \centering
     \begin{subfigure}[b]{0.45\textwidth}
         \centering
        \includegraphics[width=\linewidth]{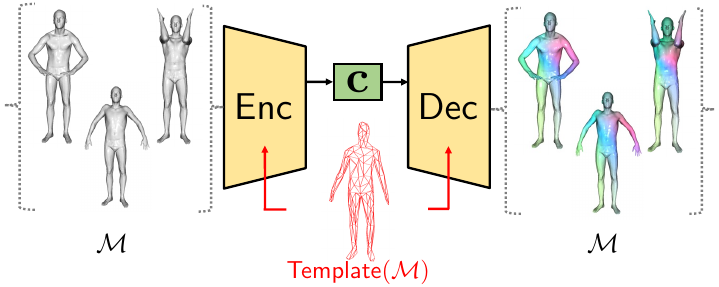}
        \caption{Prior mesh autoencoders.}
        \label{fig:intro_template}
     \end{subfigure}
     \begin{subfigure}[b]{0.45\textwidth}
         \centering
        \includegraphics[width=\linewidth]{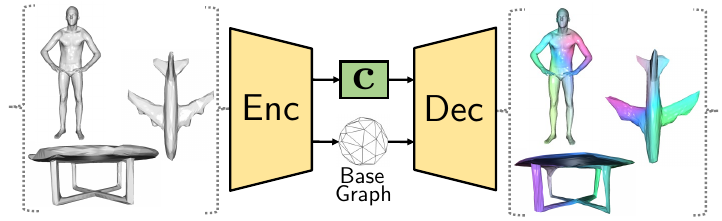}
        \caption{WrappingNet.}
        \label{fig:intro_wrapping}
     \end{subfigure}
     \caption{Prior mesh autoencoders (a) are specific to templates shared across homogeneous meshes $\mathcal{M}$. WrappingNet (b) has no such restrictions, operating on meshes of varying size/connectivity.}
     \label{fig:intro}
\end{figure}

Triangle mesh data are composed of vertex positions (list of 3D points) and their connectivity (list of vertex-index triplets). 
While previous works have successfully defined convolutions and hierarchical sampling that can deal with the non-Euclidean nature of meshes, they are limited in the types of meshes they can process. Many previous mesh autoencoders such as CoMA and its variants \cite{COMA, neural3dmm, 9150750, zhou2020fully} require mesh dataset \textit{homogeneity}, in that only the vertex properties (\eg, 3{D} positions) vary from mesh-to-mesh, but the size and connectivity remain the same (\ie, they are all mapped to a shared template mesh). 

As a result, these methods cannot be trained or evaluated on \textit{heterogeneous mesh datasets}\footnote{We use heterogeneous meshes and template-free meshes interchangeably to describe meshes that vary in connectivity and size.}, where individual meshes are variable in size and connectivity. This limitation stems from the autoencoder design having the shared template built-in to the architecture, in order to generate mesh connectivity at the decoder or align latent feature maps across meshes for inference (Fig.~\ref{fig:intro_template}). In comparison, the proposed WrappingNet (Fig.~\ref{fig:intro_wrapping}) architecture is not fixed to any particular template, allowing training and inference on heterogeneous meshes. While the mesh homogeneity requirement in prior works is sufficient for single-object tasks such as face or body animations, it prohibits the ability to perform autoencoder tasks for more general learning scenarios that are not restricted to single objects. 

In general, there are two requirements that are sufficient for a mesh autoencoder to perform both training and inference on heterogeneous mesh datasets: 
\begin{enumerate}[(RE1)]
    \item The encoder and decoder should not have mesh connectivity built-in to the architecture.
    \item The latent feature space shall be comparable across meshes of different connectivity.
\end{enumerate}
(RE1) is necessary since the model cannot perform a forward pass if the mesh's connectivity is different to the one built-in to the architecture. Some works (explained in the next section) satisfy (RE1) but fail (RE2). In these cases, the autoencoder may perform forward passes on heterogeneous meshes, but the latent space cannot be aligned and fails to be meaningful on meshes that do not share the same connectivity, preventing downstream tasks such as classification from latent codes, or latent interpolation across such meshes.

In what follows, we describe the proposed \textit{WrappingNet}, a mesh autoencoder framework that simultaneously satisfies (RE1) and (RE2). To achieve this, the latent code can be generated via any template-independent mesh encoding architecture with global pooling, which ensures a single fixed-length codeword is transmitted for \textit{any} input mesh. The fixed-length codewords enable any two mesh's codewords to exist in the same feature space. The decoder's architecture also needs to be template-independent. However, it is challenging to rebuild variable connectivity in a mesh solely from a latent code. As a result, rather than fix connectivity information into the decoder architecture itself, the encoder additionally transmits a base graph $G$, which defines the reconstructed mesh's graph topology. The decoder uses the latent code to \emph{wrap} $G$ into the reconstructed mesh.

Main contributions of this paper are the following:
\begin{enumerate}[(i)]
\item We propose WrappingNet, the first mesh autoencoder framework able to process \emph{heterogeneous meshes} yet extract a \emph{shared latent space}. To achieve this, WrappingNet uses a novel bottleneck interface consisting of a latent code and a base graph.\
\item WrappingNet is supported by novel Wrapping modules, which perform mesh deformation, and are equivariant to permutations of the mesh. These modules facilitate both the reconstruction of the mesh as well as learning descriptive latent codes.
\item Experiments show the superiority of the learned latent code on heterogeneous mesh datasets. This is verified in various tasks such as object classification from latent codes, reconstruction, and latent space analysis such as latent interpolation. 
\end{enumerate}

\section{Related Work}
\label{sec:related}
\begin{figure*}[ht!]
    \centering
    \includegraphics[width=0.75\linewidth]{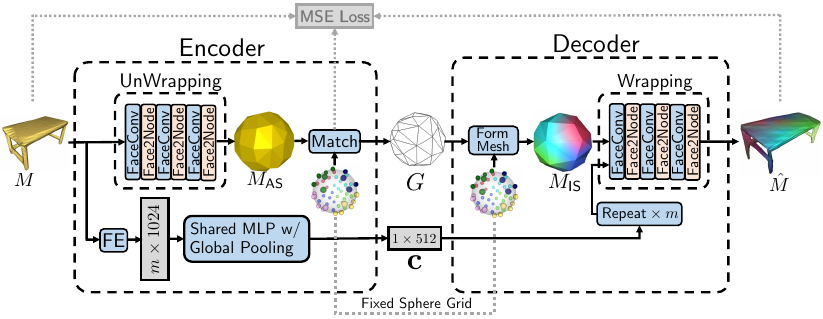}
    \caption{\textbf{WrappingNet} architecture. \textit{Encoder}: The bottom branch extracts features on the mesh faces via the feature extractor (FE) and pools them into codeword $\cv$. In the top branch, the mesh is \emph{unwrapped} into a canonical sphere mesh whose connectivity (base graph $G$) is also sent to the decoder. \textit{Decoder}: The canonical sphere mesh is recovered from the base graph and sphere grid. The copied codeword is used to \textit{wrap} the canonical sphere into the reconstructed mesh.}
    \label{fig:E2E}
\end{figure*}

\subsubsection{Mesh Autoencoders}
The first mesh autoencoder attempts can be traced back to \cite{litany2018deformable} and CoMA \cite{COMA}. The former work treats the mesh purely as a graph and applies a variational graph autoencoder (VGAE) \cite{vgae}. Since the weights of a VGAE are agnostic to the graph, this approach satisfies (RE1). However, since VGAEs extract latents on the nodes of the input graph, feature spaces can only be aligned across meshes of the same connectivity and vertex ordering, failing to satisfy (RE2). On the other hand, CoMA defines hierarchical pooling based on quadric error simplification, combined with spectral graph convolution layers, but each model instance must operate on homogeneous meshes of the same size and connectivity. This is because the graph pooling and unpooling operations are pre-computed based on a fixed template graph shared across all meshes in a homogeneous dataset. Follow-up works \cite{neural3dmm, 9150750, tretschk2020demea, ma2020cape, zhou2020fully} improve the convolution layers, latent space analysis, or applications but are still limited by the constraint of template-specific meshes. The latter work defines template-free convolutions to construct an autoencoder, but still require pre-computing the connectivity sampling based on a shared template \cite[Sec.~6]{zhou2020fully}. All these CoMA-derived methods fail to satisfy (RE1) and thus also fail (RE2).

A more recent line of work that attempts to overcome the template-specific restrictions is CoSMA \cite{semiregular, hahner22a}, an autoencoder which operates on subdivision meshes. Defined in \cite{subdivnet}, a subdivision mesh (also known as semiregular) has a hierarchical connectivity structure which can be easily up/down-sampled using Loop subdivision/pooling \cite{loop}. Such a mesh can be downsampled repeatedly until a base mesh is reached, at which Loop pooling can no longer be applied. By leveraging the subdivision structure, a CoSMA model instance does not depend on a shared template and can thus be trained on heterogeneous mesh datasets. However, CoSMA's latent space consists of features on the faces of the base mesh. Since the base mesh may vary across meshes (it is not a shared template), CoSMA satisfies (RE1) but fails (RE2), like the VGAE approach. 

A separate work of note is MeshCNN \cite{hanocka2019meshcnn}, which defines learnable up and downsampling modules that can adapt to different meshes of variable size, but these layers have not been demonstrated to construct a good autoencoder, but rather for supervised classification and segmentation applications. 

\subsubsection{Point Cloud Autoencoders}
One could also draw comparisons with autoencoders for point clouds. Although connectivity in meshes allows for definitive topological information about the underlying surface, the connectivity also brings additional challenges. In particular, a list of points can be permuted and still represent the same shape, but a mesh will not if the connectivity indices are not also permuted consistently with the points. State-of-the-art point cloud autoencoders such as FoldingNet \cite{foldingnet} and TearingNet \cite{pang2021tearingnet}, which can extract fixed-length latent representation on point clouds of different sizes, use this property of point clouds. The decoder achieves this by deforming a predefined 2D grid of points (with a fixed order) into a 3D point cloud, using the latent code as input. In a similar vein, our decoder in WrappingNet deforms a fixed 3D grid of points on a sphere, except that a connectivity is defined on it via the base graph $G$. To maintain a reasonable mesh topology on the sphere grid, a novel matching step takes place at the encoder during training, defined in the next section. This ensures that $G$'s permutation matches that of the sphere grid.

\section{WrappingNet Autoencoder}
\label{sec:overview}

We now provide a high-level overview and motivation of WrappingNet's modules, shown in Fig.~\ref{fig:E2E}, followed by specific module-level details. 

\subsection{Overview of Architecture}
\label{sec:overview}

We assume that the input to WrappingNet is a manifold mesh\footnote{We assume manifoldness due to the selected backbone architecture that performs feature extraction on the meshes, but it is not a restriction due to our proposed wrapping operations. It should not be difficult to extend WrappingNet to more general meshes.} represented using a list of vertex positions $\Xv \in \mathbb{R}^{n \times 3}$, and a list of triangles $\Tv \in \mathbb{N}^{m \times 3}$ which contain indices of the corresponding points. We thus consider meshes as vertex and triangle list tuples $M = (\Xv, \Tv)$. 

In order to generate a fixed-length codeword, the encoder adopts an initial feature extractor (FE) which computes features (of dimension 1024) on each of the $m$ mesh faces, which are then processed by a shared MLP and globally pooled (across the faces) into single codeword $\cv$ of dimension 512. Many architectures can represent the FE; we choose to use face convolution layers from SubdivNet \cite{subdivnet}, which require manifold meshes. So far, both (RE1) and (RE2) are satisfied, since the encoder's bottom branch always generates a fixed-length codeword, and the FE and shared MLP's weights are template-agnostic.

The challenge with such an encoder is that decoding a fixed-length codeword into a geometric object, such as a mesh, is nontrivial. Prior mesh autoencoders that use fixed-length codewords, such as CoMA and its variants, require the decoder to have the shared template graph hard-coded, so that the decoded mesh will always maintain the same connectivity (and node/face permutation) as that of the template. However, such a decoder violates (RE1). Rather than hard-code the connectivity into the decoder, we choose to send the connectivity across the bottleneck, which we refer to as the base graph $G$. 

To incorporate $G$ into the bottleneck, both the encoder and decoder maintain a fixed grid of points lying on the unit sphere. This fixed sphere grid defines the vertex positions of $G$ at the decoder, forming an initial sphere mesh $\MIS$. A learned module, denoted as Wrapping, uses the codeword $\cv$ as a conditional input to deform the sphere mesh $\MIS$ into the reconstructed mesh $\hat{M}$. The decoder, designed in this way, takes graphs whose vertices lie on the sphere grid, so that the $v$-th vertex takes the $v$-th position in the sphere grid; not all points on the sphere grid may be connected.

The remaining challenge is how to permute the mesh's vertex order at the encoder, so that the initial sphere mesh maintains a reasonable sphere-like shape. Due to the fixed ordering of the sphere grid, the input mesh's vertex ordering may not be aligned with that of the sphere grid. To solve this, the encoder's UnWrapping module is trained to deform the input mesh into an approximate sphere mesh $\MAS$, whose positions are ``projected'' onto the sphere grid via a matching step. The matching assigns each vertex in $\MAS$ a unique point on the sphere grid, allowing the encoder to form $G$ by permuting the vertex ordering of $\MAS$'s connectivity via the assignment. This assignment also allows for a vertex-to-vertex mean square error (MSE) loss to be computed between the reconstructed mesh $\hat{M}$ (whose vertex ordering is that of the sphere grid) and the input mesh $M$. 

To train the model end-to-end, this MSE loss between $M$ and $\hat{M}$ is used to train the Wrapping module in the decoder, as well as the FE and shared MLP layers at the encoder. The UnWrapping module at the encoder is supervised with a Chamfer loss between the positions of $\MAS$ and randomly sampled points on the unit sphere, which helps ensure that $\MAS$ is an approximate sphere mesh. 

\subsubsection{Motivation of Sphere Grid}
When designing the decoder to make use of input connectivity $G$, initializing $G$'s vertex positions to be on the sphere is a natural solution. Other ideas bring additional challenges. One could send the entire approximate sphere mesh $\MAS$ across the bottleneck, which would avoid the need for the sphere grid or matching. However, this may cause entanglement with the codeword $\cv$ regarding the global shape of the input object, which we want to be encoded entirely by $\cv$ for maximal descriptiveness. This is because $\MAS$'s vertex positions do not lie perfectly on the sphere (since the UnWrapping is not perfect), and may still contain information about the original mesh's shape. Using the sphere grid to enforce a sphere geometry for $\MIS$ at the decoder ensures that $\cv$ maximally encodes the object shape, since a sphere grid is generic. This is later verified in an ablation study. 

By deforming a sphere grid, our approach resembles TearingNet \cite{pang2021tearingnet}, a point cloud autoencoder whose decoder uses a codeword to deform a fixed 2D grid which can also have edge connectivity, yielding a mesh at the output. However, the differences are that WrappingNet (i) uses the exact graph topology $G$ sent from the encoder, (ii) uses mesh-based feature processing, and (iii) enforces a vertex-to-vertex loss, rather than Chamfer loss. The latter point is only possible due to the matching step between the input points and the fixed grid. These differences allow WrappingNet to achieve superior surface reconstruction performance over TearingNet, as will be shown in the results.

In summary, WrappingNet satisfies (RE1) since the encoder and decoder are not specific to one fixed connectivity, and satisfies (RE2) due to the global feature pooling at the encoder. In the next section, we provide further details and motivation behind the wrapping operations, sphere grid, and matching step.

\subsection{WrappingNet Modules}

\subsubsection{Wrapping Operations}
\label{sec:deformation_backbone}

We now describe our novel wrapping modules, which are used in the UnWrapping step in the encoder and the Wrapping step in the decoder. These modules can be thought of as learnable mesh deformation layers, since they adjust the vertex positions of the mesh, but do not change its connectivity. Shown in Fig.~\ref{fig:E2E}, their architectures are comprised of face convolution (denoted FaceConv) layers from \cite{subdivnet}, and our novel \textit{Face2Node} layers, which are alternated. At a high level, the FaceConv layers operate on face-wise features, but in order to deform a mesh, we need to update the vertex positions, which are node-wise features. This necessitates a face to node conversion. As mentioned previously, the use of face convolution layers from \cite{subdivnet} requires manifoldness, but this can be adjusted in future work.

The input face features used for these modules are deterministic features extracted from the meshes $M$ and $\MIS$. They are chosen to be features in local rather than global coordinates, which improves the performance of these layers (full details in supplementary). At the decoder, these features from $\MIS$ are each concatenated with the codeword $\cv$. 

\subsubsection{Face2Node} 
The Face2Node layer consumes a mesh containing face features $\Fv$ and (i) predicts differential vertex position updates, and (ii) optionally updates the face features. 

\begin{figure}
    \centering
    \includegraphics[width=0.215\linewidth]{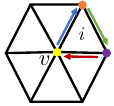}
    \caption{Concatenation order in \eqref{eq:concat_order}. In this example, the edge vectors are concatenated in the blue, green, red order. For node indices; yellow: $j(v)$; orange: $j(v)+1 \mod 3$; purple: $j(v)+2 \mod 3$.}
    \label{fig:Face2NodeOperational}
\end{figure}

Let $v$ be a node we are updating, $\mathcal{N}_v$ denote the faces that node $v$ is a part of, and let $\Fv_i$ denote the $i$-th face's feature, and suppose $i \in \mathcal{N}_v$, \ie, $v$ is a node of face $i$. Face2Node first creates a new feature for face $i$ relative to $v$ by concatenating edge vectors of face $i$ to $\Fv_i$, \ie, if $\xv\ssi{0}, \xv\ssi{1}, \xv\ssi{2}$ are the three vertex positions of face $i$, its edge vectors are $\ev\ssi{k} = \xv\ssi{k+1 \mod 3}-\xv\ssi{k}$, $k \in \{0,1,2\}$. We ensure the edge vectors are concatenated in an order-equivariant manner, \ie, if the local node ordering \textit{within a face} is permuted, the concatenation order permutes accordingly. Since nodes within a face are ordered so that normals point outward, this can be achieved by fixing a relative starting index. We set the starting index to be the node index within face $i$ that $v$ happens to be, denoted as $j(v)$. The concatenated feature for face $i$ relative to $v$ is given by
\begin{equation}
    \fv_i\ssi{j(v)} = \Fv_i || \ev\ssi{j(v)}||\ev\ssi{j(v)+1 \mod 3}||\ev\ssi{j(v)+2 \mod 3}
    \label{eq:concat_order}
\end{equation}
where $||$ denotes concatenation. A figure illustrating this process can be found in Fig.~\ref{fig:Face2NodeOperational}. 

\begin{figure}
    \centering
    \includegraphics[width=\linewidth]{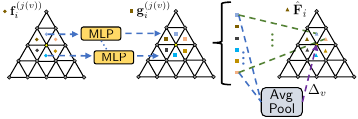}
    \caption{\textbf{Face2Node} module. Node $v$: yellow node. $\mathcal{N}_v$: faces with colored dots. }
    \label{fig:Face2Node}
\end{figure}

As shown in Fig.~\ref{fig:Face2Node}, Face2Node then updates the concatenated face features (for all faces in $\mathcal{N}_v$) using a shared MLP, \ie, $\gv_i\ssi{j(v)} = \mathrm{MLP}(\fv_i\ssi{j(v)})$. The differential position update for node $v$ is the average of the first 3 components of $\gv_i\ssi{j(v)}$, averaged over $\mathcal{N}_v$. 

Then the differential position update of $v$ is $\Delta_v = \frac{1}{|\mathcal{N}_v|} \sum_{i \in \mathcal{N}_v}  \gv_i\ssi{j(v)}\text{[0:3]}$, and
the node position is updated as $\xv_v \leftarrow \xv_v + \Delta_v$. 

The face features are then updated by averaging over all three versions of the remaining entries $\gv_i\ssi{j(v)}\text{[3:]}$ for the $i$-th face, $\hat{\Fv}_i \leftarrow \frac{1}{3}(\gv_i\ssi{0}\text{[3:]}+\gv_i\ssi{1}\text{[3:]}+\gv_i\ssi{2}\text{[3:]})$,
which ensures the features on the face remain invariant to the order of its nodes. This is important because the FaceConv layers are designed to be agnostic to specific node orderings.

\subsubsection{Sphere Grid}
The sphere grid is defined by the Fibonacci lattice \cite{fibonaccilattice}. For the grid, the $i$-th point (out of $N$ total points) has 3D coordinates given by 
\begin{align}
    \bar{x}_i &=  \sqrt{1 - (1-(i/(N-1))^2)^2} \cos (\phi i), \nonumber \\
    \bar{y}_i &= 1-(i/(N-1))^2,  \label{eq:sphere_grid} \\
    \bar{z}_i &= \sqrt{1 - (1-(i/(N-1))^2)^2} \sin (\phi i), \nonumber
\end{align}
where $\phi \triangleq \pi (3 - \sqrt{5})$ is the Golden angle. We set $N$ large enough to ensure that $N$ is larger than the maximum number of vertices of any mesh to be encountered. 

\subsubsection{Matching}
The sphere grid coordinates in \eqref{eq:sphere_grid} are defined in a fixed, deterministic order, which is not guaranteed to be aligned with the vertex order of $\MAS$. As a result, we compute a mapping between the vertex positions of $\MAS$ and sphere grid coordinates by solving an assignment problem, with Euclidean distance as the pairwise cost, via the Hungarian algorithm. The matching step ensures that $G$ is vertex permutation invariant, since its vertex order is that of \eqref{eq:sphere_grid}. This is aided by the fact that the Wrapping module is equivariant to vertex permutation, ensuring a consistent matching no matter the input permutation. Further details of the matching step can be found in the supplementary.   

In summary, all meshes in Fig.~\ref{fig:E2E} ($M$, $\MAS$, $\MIS$ and $\hat{M}$) use the same connectivity, which is given from the input $M$. However, $M$ and $\MAS$ share the same input vertex permutation, whereas $\MIS$ and $\hat{M}$ at the decoder (as well as $G$) share the vertex permutation of the sphere grid \eqref{eq:sphere_grid}. The matching step allows a correspondence between these two orderings to be computed.


\section{Experimental Results}
\label{sec:results}

We focus on heterogeneous mesh datasets containing meshes of varying categories, with no shared templates. The datasets chosen all contain 2-manifold meshes. In our experiments, we demonstrate WrappingNet on these datasets using the architecture in Fig.~\ref{fig:E2E}, which we refer to as vanilla WrappingNet. Since the chosen manifold datasets are of low density ($\approx$ 250 vertices each), we additionally use the remeshed versions of the datasets as in \cite{subdivnet}, which possess subdivision structure at higher density ($\approx 3000$ vertices). We then use a subdivision version of WrappingNet (full architecture in Fig.~V of appendix) that operates on subdivision meshes, which allows for stronger hierarchical feature extraction to reflect finer details. Note that any 2-manifold mesh can be remeshed into a subdivision mesh \cite{neuralsubdiv}, so it is not difficult to have subdivision meshes once the manifold requirement is satisfied.

Since subdivision meshes have a fixed indexing relative to the simplified \textit{base mesh} of the subdivision hierarchy, it is straightforward to incorporate Loop pooling/subdivision into the WrappingNet framework. At the encoder, Loop pooling of faces is applied within the feature extractor (FE) module; at the decoder, Loop subdivision is performed after the Wrapping module. The wrapping operations now operate at the (pooled) base mesh, rather than the input mesh. One plausible mesh autoencoder that satisfies (RE1) and (RE2) with subdivision is to transmit the base mesh (containing vertex positions) rather than base graph. However, this model would learn a latent code merely for super-resolution of the base mesh, and fail to represent the object shape, which would be contained in the base mesh; see Sec.~\ref{sec:ablation} ~ for an ablation study demonstrating this.

\subsection{Experimental Setup}
\label{sec:experiments:setup}
\textbf{Heterogeneous Mesh Datasets}:
We use SHREC11 \cite{SHREC11} and Manifold40 \cite{subdivnet}. None of the meshes are guaranteed to have the same connectivity, even within the same category. Manifold40 is a manifold version of ModelNet40 \cite{modelnet40}. To generate the subdivision version of the datasets, we use the method in \cite{neuralsubdiv}, which generates a simplified base mesh, subdivides it, and projects vertices back to the original mesh; we use 3 subdivision levels. For both datasets, we normalize the vertex positions to be zero-mean and within the unit sphere, and use default train-test splits.

\textbf{Implementation Details}:
WrappingNet is trained by enforcing the vertex-to-vertex MSE loss between $M$ and $\hat{M}$, with backpropagation through the decoder and bottom branch of the encoder. We detach the gradient at $\MAS$. As mentioned in Sec.~\ref{sec:overview}, the encoder's UnWrapping module output is supervised using a Chamfer loss against samples on a unit sphere. This is done since we want the decoder to reconstruct $M$'s shape from $\cv$, not from $G$, which should only transmit the graph topology. We use a latent code of length 512, and Adam \cite{ADAM} with a learning rate of $5 \times 10^{-5}$. We note that both models have low computational complexity due to the FaceConv layers (full details in supplementary).

We compare with point cloud autoencoders FoldingNet \cite{foldingnet} and TearingNet \cite{pang2021tearingnet} that extract fixed-length latent codes from variable-size point clouds. We trained them with latent codes of size 512, on vertex positions of the subdivided meshes, since they are tuned for point clouds of similar density.  

\begin{table}[t]
  \centering\tiny
  \caption{Average classification metrics from test latent codes using SVM with 5-fold CV. Standard deviations over CV are $\leq 0.015$.}
    \begin{tabular}{c||ccc|ccc}
    \hline
    Dataset  & \multicolumn{3}{c|}{Manifold10} & \multicolumn{3}{c}{Mainfold40} \bigstrut\\
    \hline
    Metric & Acc.  & Prec. & Recall & Acc.  & Prec. & Recall \bigstrut\\
    \hline
    \hline
    FoldingNet &  91.0\% & 0.901 & 0.893 & 82.5\% & 0.756 & 0.678 \bigstrut[t]\\
    TearingNet &  90.8\% & 0.901 & 0.892 & 82.9\% & 0.750 & 0.664 \\
    WrappingNet (vanilla) & 90.8\% & 0.901 & 0.891 & 83.0\% & 0.760 & 0.710 \\
    WrappingNet (subdiv.) & 91.0\% & 0.908 & 0.902 & 83.3\% & 0.778 & 0.730 \bigstrut[b]\\
    \hline
    \end{tabular}%
  \label{tab:class_results}%
  \vspace{-10pt}
\end{table}%

\begin{figure}[t]
    \centering
    \includegraphics[width=0.55\linewidth]{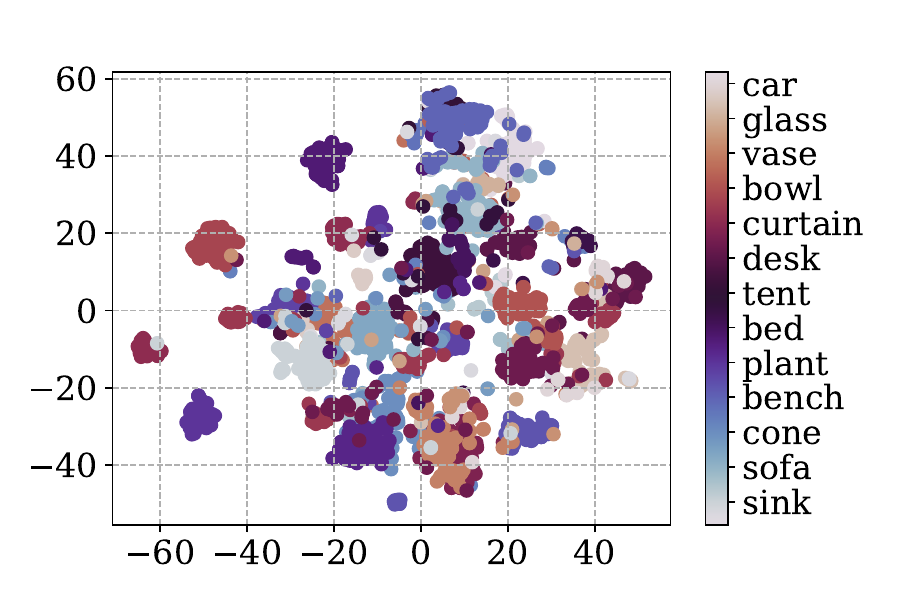}
    \caption{t-SNE plot of WrappingNet latent space, Manifold40. }
    \label{fig:tsne}
    \vspace{-10pt}
\end{figure}

\subsection{Latent Code For Shape Representation}

\textbf{Shape Classification Task}:
One common method to examine a learned latent space of 3D objects is to examine how well they can be used to classify shape class. We first evaluate the classification accuracy on test latent codes from SHREC11, Manifold40, and Manifold10 (a subset of Manifold40). Specifically, we first train WrappingNet, extract test latent codes, and train an SVM classifier on them with 5-fold cross-validation (CV). SVM classification is chosen in order to directly classify in the latent space. Note that it is only possible to perform this experiment on such mesh data with WrappingNet since it extracts a shared latent space across meshes, unlike prior mesh autoencoders. 

We report the accuracy and macro-averaged precision and recall in Tab.~\ref{tab:class_results}. WrappingNet performs competitively with point cloud autoencoders, demonstrating the learned latent spaces are of comparable quality in distinguishing objects. This is expected since point clouds and meshes are two representations of the same shape; high-level information such as object class should not be fragile to the representation at dense sampling rates. Similarly, the subdivision WrappingNet helps improve performance, but not significantly. We omit the classification results on SHREC11 here since all models achieved nearly perfect classification scores on SHREC11 ($>0.99$ for all metrics).

\textbf{t-SNE Visualization}:
We also visualize the t-SNE \cite{tsne} plots of the test latent codes extracted from subdivision WrappingNet for the Manifold40 in Fig.~\ref{fig:tsne}. As can be seen, the latent codes of differing classes are clustered by class, demonstrating that the shape information has been encoded into the latent code; the figure appears identically for vanilla WrappingNet. 

\subsection{Reconstructed Shape Analysis}

\begin{table}[t]
  \centering\scriptsize
  \caption{Topology comparison of reconstructed shapes. Both a point cloud and mesh rendering are provided for each example.}
    \begin{tabular}{c|c||ccc}
    \hline
          &       & Ground Truth    & TearingNet & WrappingNet \bigstrut\\
    \hline
    \multirow{2}[2]{*}{\begin{sideways}Example 1\end{sideways}} & \begin{sideways}\phantom{iiiiii}Mesh\phantom{ii}\end{sideways} & \includegraphics[width=0.18\linewidth]{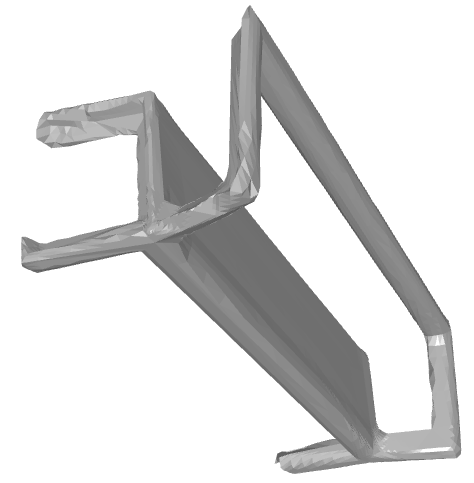} & \includegraphics[width=0.18\linewidth]{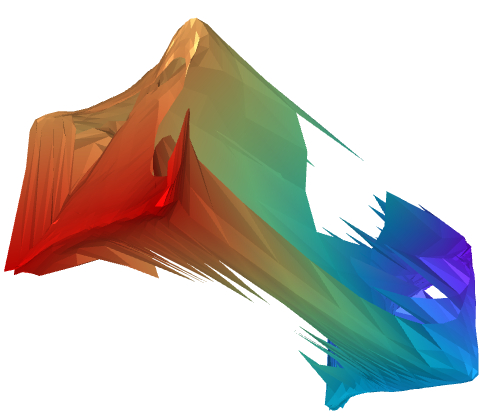} & \includegraphics[width=0.18\linewidth]{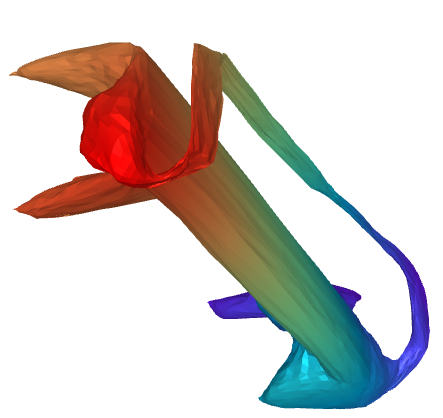} \bigstrut[t] \vspace{-.5em} \\
          & \begin{sideways}Point Cloud\end{sideways} & \includegraphics[width=0.18\linewidth]{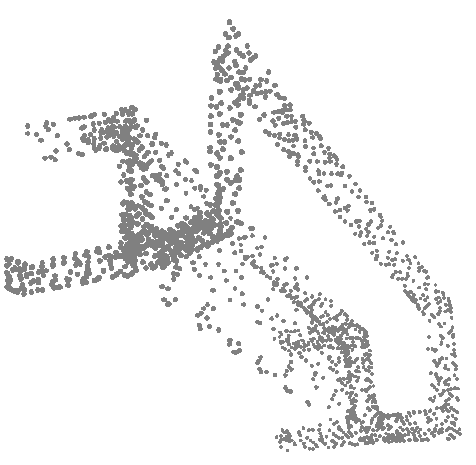} & \includegraphics[width=0.18\linewidth]{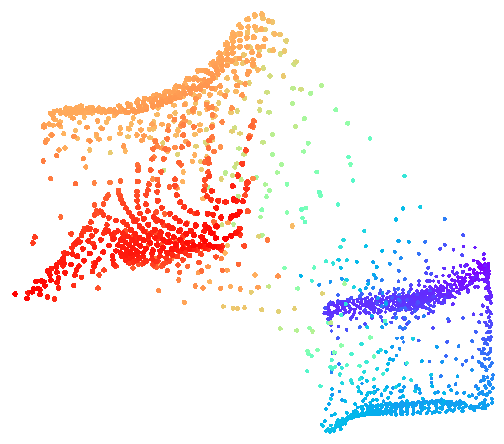} & \includegraphics[width=0.18\linewidth]{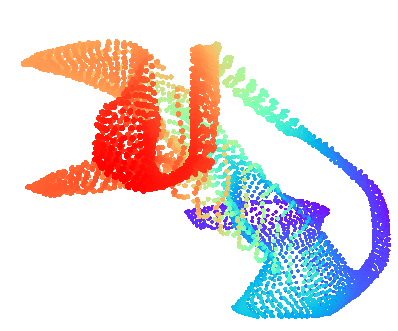} \bigstrut[b]\\
    \hline
    \hline
    \multirow{2}[2]{*}{\begin{sideways}Example 2\end{sideways}} & \begin{sideways}\phantom{iiiiii}Mesh\phantom{iii}\end{sideways} & \includegraphics[width=0.15\linewidth]{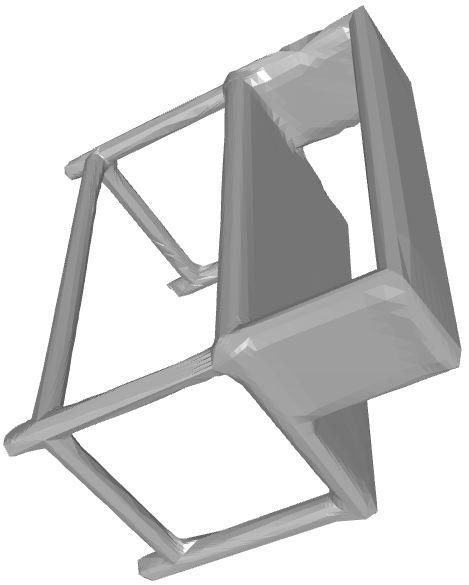} & \includegraphics[width=0.15\linewidth]{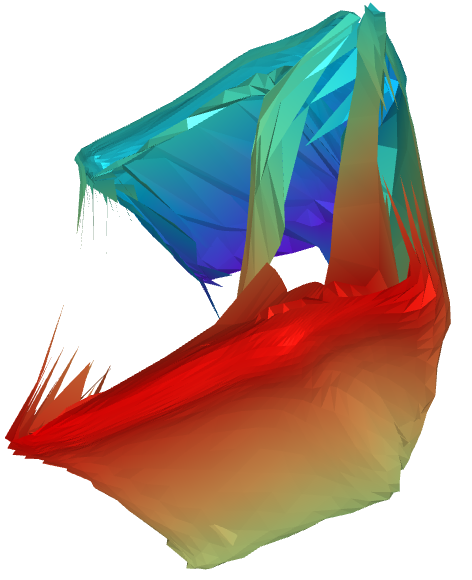} & \includegraphics[width=0.15\linewidth]{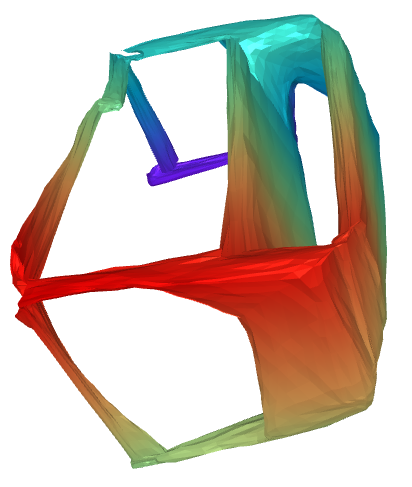} \bigstrut[t]  \vspace{-.4em}  \\
          & \begin{sideways}Point Cloud\end{sideways} & \includegraphics[width=0.15\linewidth]{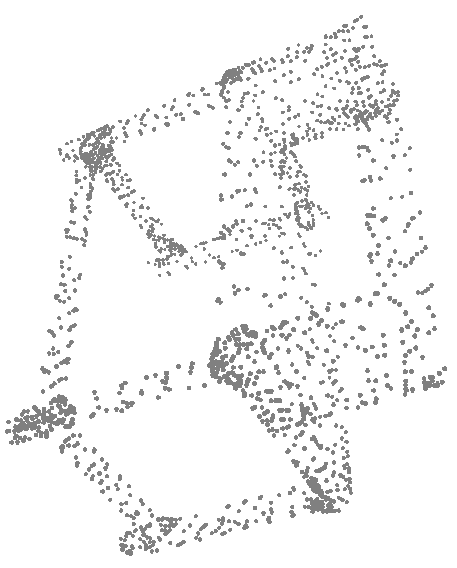} & \includegraphics[width=0.15\linewidth]{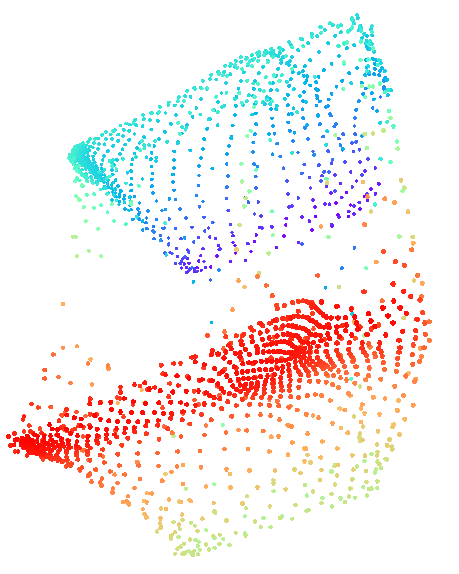} & \includegraphics[width=0.15\linewidth]{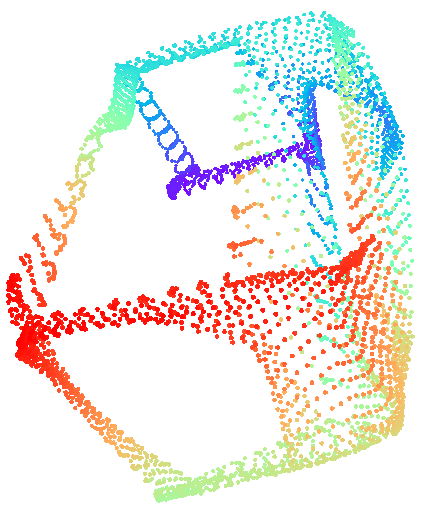} \bigstrut[b]\\
    \hline
    \end{tabular}%
  \label{tab:topology_comparison}%
  \vspace{-1em}
\end{table}%

We demonstrate new capabilities achieved in shape reconstruction using WrappingNet. We focus on the subdivision version of the mesh datasets due to finer reconstruction details and fair comparison with point cloud autoencoders. Extended results can be found in supplementary.

\textbf{Improved Reconstruction Topology}: We first qualitatively compare the topology of reconstructed shapes between WrappingNet, FoldingNet, and TearingNet, shown in Tab.~\ref{tab:topology_comparison}. Since point clouds lack connectivity between points, the topological information of the underlying shape is lost; however, this is not the case in meshes. We see that TearingNet (a more topology-friendly version of FoldingNet) introduces points under the bench's front and side that do not exist in the ground truth shape. The inferred meshes from TearingNet also inherit such errors from the point geometry, and do not preserve the topology. In comparison, WrappingNet is aided by mesh connectivity via the base graph and preserves the original topology.

\begin{table}[!t]
  \centering\tiny 
    \caption{Reconstruction performance in terms of chamfer distance (CD), normals error (NE), and curvature preservation (CP).}
    \begin{tabular}{c|ccc}
    \hline
    Dataset & SHREC11 & Manifold10 & Mainfold40 \\
    Metrics & CD / NE / CP & CD / NE / CP & CD / NE / CP \\
    \hline
    \hline
    FoldingNet         & \textbf{0.013} / 0.12 / 0.021          & \textbf{0.001} / 0.10 / 0.004                   & 0.006 / 0.28 / 0.011 \\
    TearingNet & \textbf{0.013} / 0.11 / 0.013          & \textbf{0.001} / 0.09 / 0.003          & \textbf{0.005} / 0.26 / 0.009 \\
    \textbf{WrappingNet} & 0.023 / \textbf{0.09} / \textbf{0.004} & \textbf{0.001} / \textbf{0.07} / \textbf{0.001} & \textbf{0.005} / \textbf{0.22} / \textbf{0.004} \\
    \hline
    \end{tabular}%
  \label{tab:chamfer_results}%
  \vspace{-10pt}
\end{table}%

We additionally compare several reconstruction metrics to further demonstrate the superior topology learned by WrappingNet. Shown in Tab.~\ref{tab:chamfer_results}, we evaluate point-based reconstruction metrics for comparison with point cloud autoencoders. Specifically, we use Chamfer distance (CD), normals error (NE), and curvature preservation (CP) \cite{potamias2022eccv} (full details in supplementary). 
WrappingNet significantly outperforms FoldingNet and TearningNet in NE and CP. The additional connectivity helps to better preserve local surface information, as it guides the mesh deformation during reconstruction and enables sphere matching for the vertex-to-vertex training loss. Our method also shows similar or better performance in CD of the Manifold datasets, but other methods show better results for SHREC11. This is reasonable considering that FoldingNet and TearingNet use CD as a training objective. Additionally,
as CD measures unordered point-to-point distances, a lower CD does not necessarily correlate with better surface preservation, whereas the vertex-to-vertex loss does.

\begin{figure}[t]
    \centering
     \begin{subfigure}[b]{0.5\textwidth}
         \centering
        \includegraphics[width=0.8\linewidth]{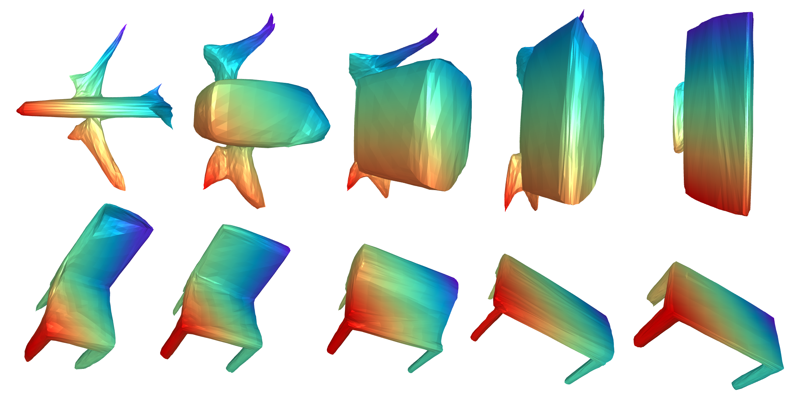} \vspace{-.3em}
        \caption{\textit{Inter}-class. $1^{\text{st}}$ row: airplane $\rightarrow$ monitor. $2^{\text{nd}}$ row: chair $\rightarrow$ table.}
     \end{subfigure}
      \begin{subfigure}[b]{0.5\textwidth}
         \centering
        \includegraphics[width=0.8\linewidth]{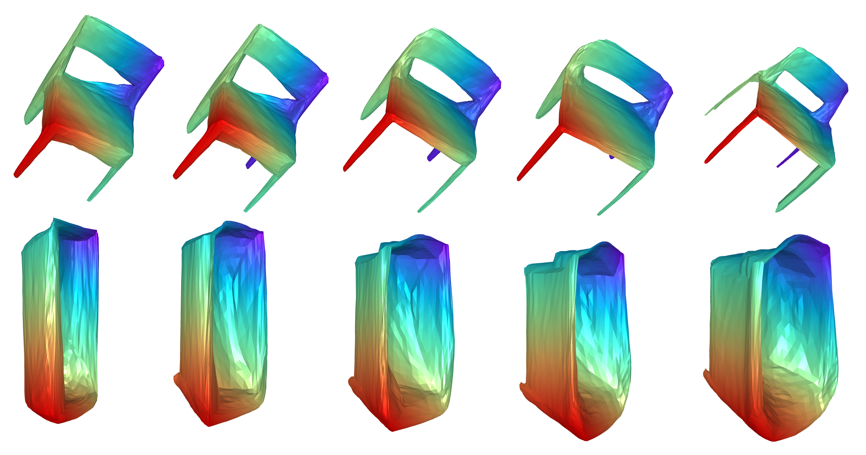} \vspace{-.3em}
        \caption{\textit{Intra}-class. $1^{\text{st}}$ row: chair. $2^{\text{nd}}$ row: bathtub.}
     \end{subfigure}
     \caption{Latent space interpolation. Base graph used comes from the encoder output of the ground truth mesh of the left side. }
     \label{fig:latent_interp}
\end{figure}

\textbf{Latent Interpolation}: WrappingNet is also the first mesh autoencoder able to perform latent space interpolation on heterogeneous mesh datasets, shown in Fig.~\ref{fig:latent_interp}; this is only possible by satisfying both (RE1) and (RE2). This can be done by interpolating two latent codes from two separate meshes, while fixing the base graph to be from one of the two meshes. As shown, interpolated latent codes generate intermediate shapes, demonstrating that the latent code encodes the object shape.

\textbf{Reconstruction from Varying Base Graphs}:
WrappingNet's bottleneck consists of both the base graph and latent code. Rather than decode varying latent codes with a fixed base graph (as in latent interpolation), one can also decode a varying base graph with a fixed latent code. The base graph may contain a different graph topology than the object represented by the latent code. In Tab.~\ref{tab:mix_match} (left most column), we show three Manifold40 meshes; $M_1$ (top) and $M_2$ (middle) are \textbf{genus-2}, while $M_3$ (bottom) is \textbf{genus-0}. We use the encoder to extract latent codes $\cv_i$ and base graphs $G_i$ from each of these meshes, \ie, $\cv_i, G_i \leftarrow \text{Encoder}(M_i) , \text{where } i \in \{1,2,3\}$. Tab.~\ref{tab:mix_match} shows decoded meshes from all latent code-base graph combinations.

We see that when the genus of the graph topology matches (\eg, between $M_1$ \& $M_2$), the decoder outputs meshes resembling the shapes given by the input latent code; whether the two holes in $M_1$ and $M_2$ exist on the chair legs or armrests is determined by the latent code. When the genus is mismatched (\eg, between $M_1$ \& $M_3$), the decoder tries its best to wrap the wrong genus base graph into the shape defined by the codeword. This is especially visible with $\text{Decoder}(\cv_1, G_3)$ (first row, last column) resembling the chair in $M_1$ but missing holes under the armrests, and $\text{Decoder}(\cv_3, G_2)$ (last row, second column) resembling the table in $M_3$ but having holes on its sides.
These results demonstrate that: (i) the global shape information is contained primarily in the codeword (as intended by design), (ii) the base graph carries more impact on the graph topology, and (iii) as long as the genus of the base graph matches that of the input mesh, WrappingNet can produce a reasonable reconstructed mesh.

\subsection{Ablation Study}
\label{sec:ablation}

\begin{table}[t]
  \centering\scriptsize
  \caption{Reconstructions from various base graphs \& latent codes.}
    \begin{tabular}{rc|ccc}
    \hline
          &       & \multicolumn{3}{c}{Base graph comes from} \vspace{-.3em} \bigstrut[t]\\
          &       \phantom{\includegraphics[width=0.155\linewidth]{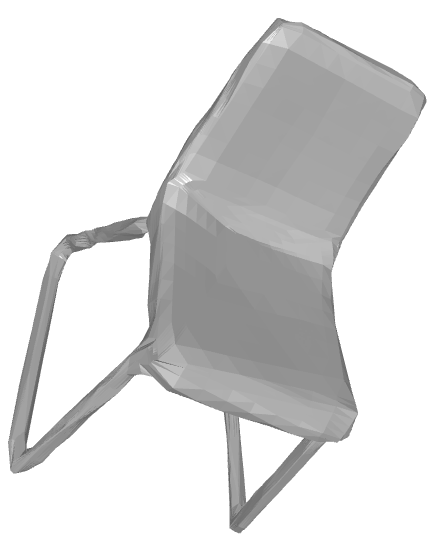}}& \includegraphics[width=0.15\linewidth]{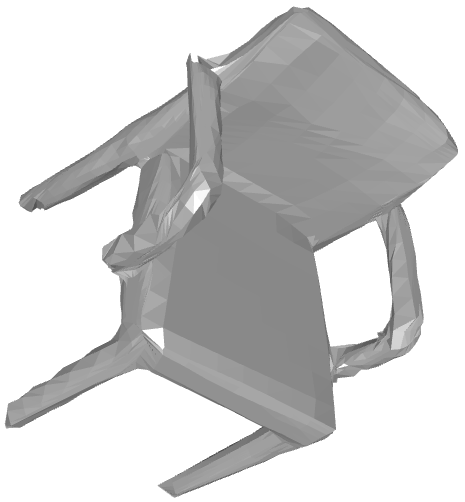}    & \includegraphics[width=0.15\linewidth]{diagrams/figures_decoder/m2_gt_gray.png}    & \includegraphics[width=0.15\linewidth]{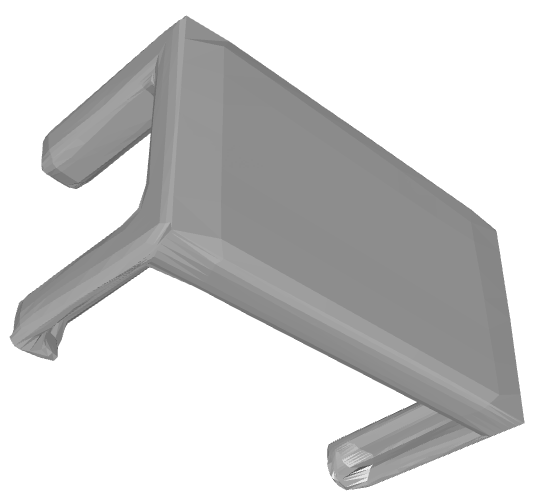} \vspace{-.3em} \bigstrut[b]\\
    \hline
    \multirow{3}[2]{*}{\begin{sideways}\phantom{aaaa}Latent codes comes from\end{sideways}} & \includegraphics[width=0.15\linewidth]{diagrams/figures_decoder/m1_gt_gray.png}   & \includegraphics[width=0.17\linewidth]{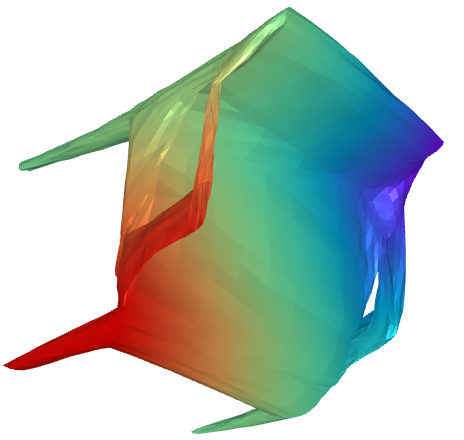}  & \includegraphics[width=0.15\linewidth]{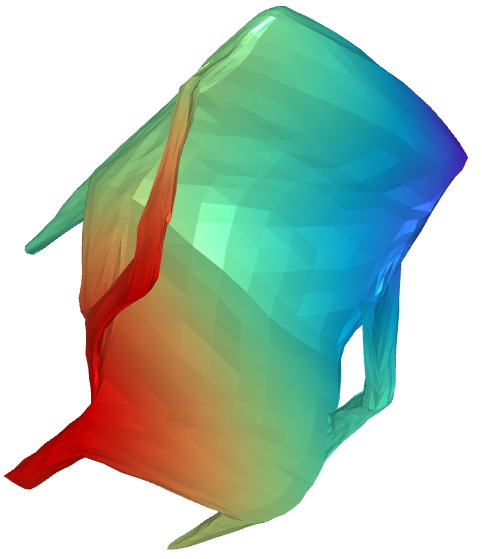}  & \includegraphics[width=0.18\linewidth]{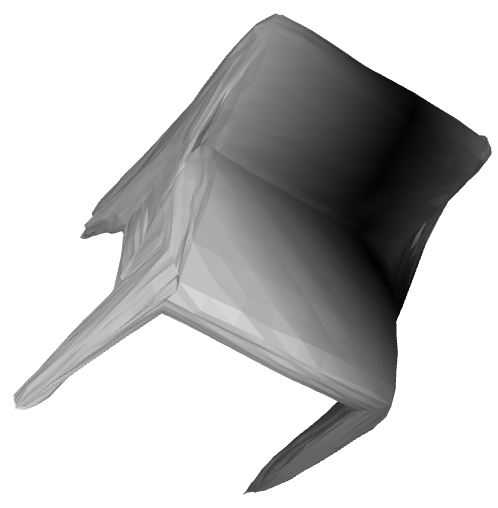} \phantom{\includegraphics[width=0.001\linewidth,height=0.17\linewidth]{diagrams/topology_comparison/2_gt_pc.png}} \bigstrut[t]\\
          & \includegraphics[width=0.15\linewidth]{diagrams/figures_decoder/m2_gt_gray.png}   & \includegraphics[width=0.18\linewidth]{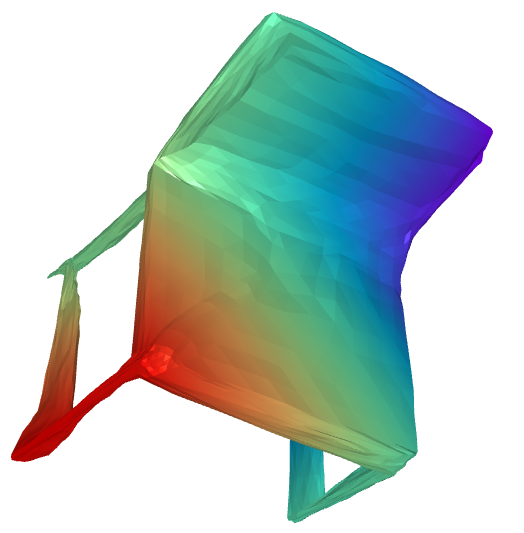}  & \includegraphics[width=0.15\linewidth]{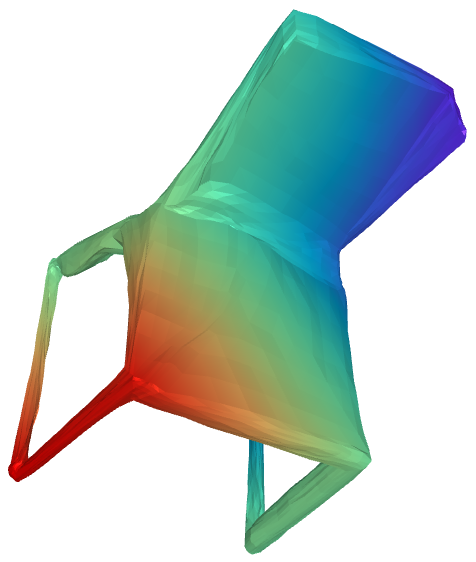}  & \includegraphics[width=0.15\linewidth]{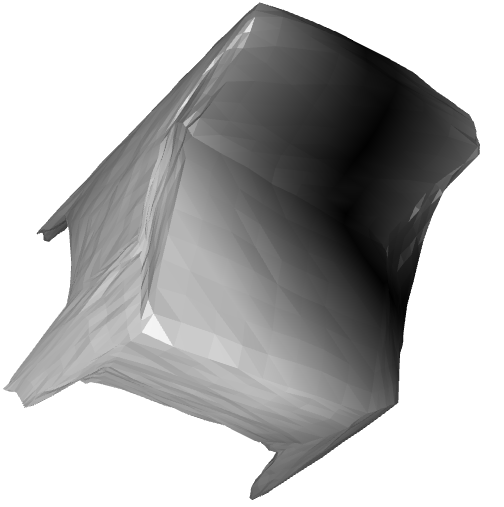} \\
          & \includegraphics[width=0.15\linewidth]{diagrams/figures_decoder/m3_gt_gray.png}   & \includegraphics[width=0.15\linewidth]{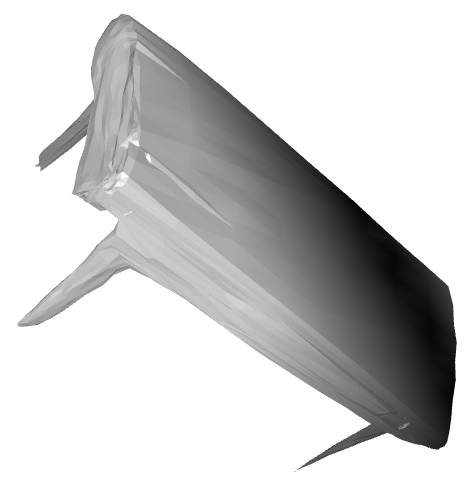}  & \includegraphics[width=0.15\linewidth]{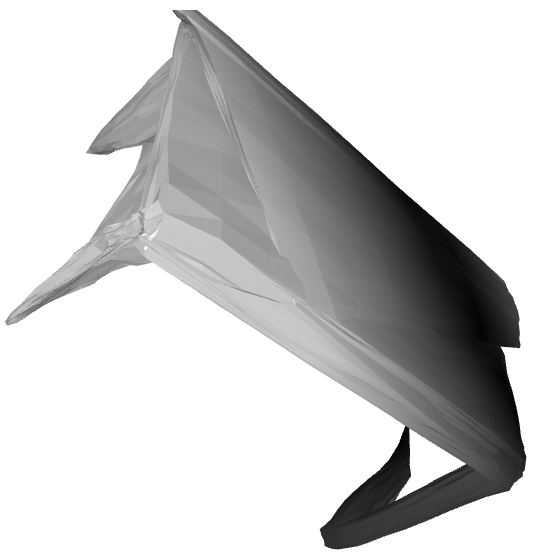}  & \includegraphics[width=0.15\linewidth]{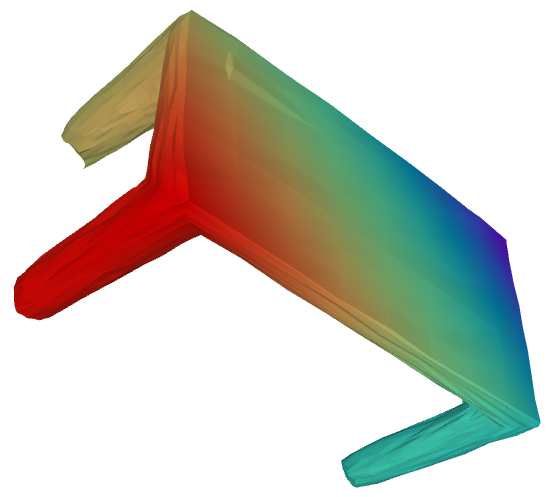} \bigstrut[b]\\
    \hline
    \end{tabular}%
  \label{tab:mix_match}%
  \vspace{-10pt}
\end{table}%

\begin{table}[t]
  \centering\tiny
  \caption{Classification (SVM on latent codes) performance of different configurations of WrappingNet on Manifold40 (ablation).}
    \begin{tabular}{l|cccc}
    \hline
    Config. & UnWrapping & Matching & Disentangle & Acc. \bigstrut\\
    \hline
    Send Base Mesh   & $\times$     & $\times$     & Least & 57.60\% \bigstrut[t]\\
    Send $\MAS$    & $\checkmark$     & $\times$     & More  & 81.20\% \\
    Send $G$    & $\checkmark$     & $\checkmark$     & Most  & \textbf{83.30\%} \bigstrut[b]\\
    \hline
    \end{tabular}%
  \label{tab:ablation}%
  \vspace{-10pt}
\end{table}%

We examine the effect of two aspects of WrappingNet that are critical for learning descriptive latent codes: (a) the sphere matching step, and (b) the use of the UnWrapping module at the encoder. As mentioned previously, without (a), the model (vanilla or subdivision) could send $\MAS$ instead of $G$. Without (b), a subdivision model could send the entire base mesh instead of $G$; both of these would satisfy (RE1) and (RE2). The decoder would replace $\MIS$ with these meshes instead. Aspect (b) pertains exclusively to subdivision WrappingNet, since if the vanilla model did not have (b), its encoder would be sending the ground truth mesh.

We examine these two configurations on the subdivision meshes, and compare with full WrappingNet (sends the base graph $G$). Note that without (b), it is implied that no sphere matching takes place. We evaluate 5-fold CV SVM classification on test latent codes
, shown in Tab.~\ref{tab:ablation}.
Without the UnWrapping (sending the base mesh), the latent code and base mesh vertex positions are \textit{entangled}. Some of the shape (and hence class) information is contained in the base mesh rather than completely in the latent code, which results in poor classification performance. The latent code is rather used to superresolve the base mesh. 
With UnWrapping but no matching (sending $\MAS$), $\MAS$ is mostly disentangled from the latent code and the decoder needs to rely more on the latent codes for reconstruction. This forces more information to be successfully ``pushed'' into the latent codes, and thus the classification accuracy increases.
Finally, the use of sphere matching allows further disentanglement 
by ensuring the vertex positions lie perfectly on the sphere, providing the best classification accuracy. t-SNE plots for the respective latent spaces can be found in supplementary.

\section{Conclusion}
\label{sec:conclusion}
Heterogeneity in mesh data due to variable connectivity and size has presented challenges in designing mesh autoencoders. WrappingNet is the first method to learn a latent space across such meshes, using a novel bottleneck interface consisting of a latent code and base graph. We demonstrate improved surface reconstruction, and the ability to extract a descriptive latent space from heterogeneous meshes.

\bibliography{ref}

\appendix

\clearpage
\section*{\centering \LARGE Appendix}
\renewcommand{\thesection}{\Roman{section}} 
\renewcommand{\thesubsection}{\thesection-\Alph{subsection}}
\renewcommand{\thetable}{\Roman{table}}
\renewcommand{\thefigure}{\Roman{figure}}
\renewcommand{\theequation}{\roman{equation}}

\setcounter{figure}{0}
\setcounter{table}{0}
\setcounter{equation}{0}

\section{Additional Experimental Results}
We first discuss additional experimental results, paralleling Sec.~4.2 (latent code for shape representation), 4.3 (reconstructed shape analysis), and 4.4 (ablation study) in the main text.

\subsection{Latent Code for Shape Representation}
We first provide additional t-SNE plots, followed by a new experiment demonstrating the added benefits of connectivity information for shape representation when the number of points used to represent the shape is low.
\subsubsection{Additional t-SNE Visualization}
\begin{figure}[t]
     \centering
     \begin{subfigure}[b]{0.48\linewidth}
         \centering
        \includegraphics[width=\linewidth]{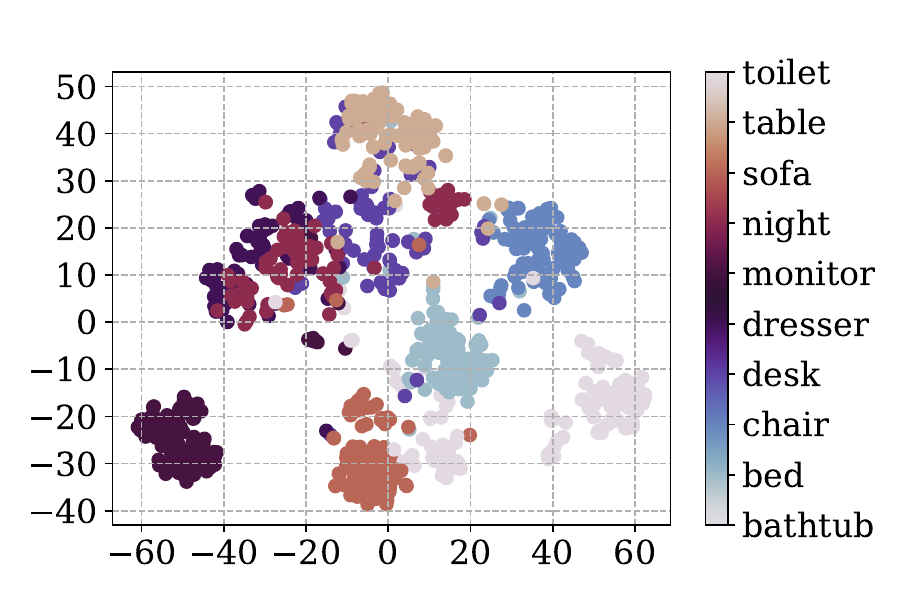}
        \caption{Manifold10.}
     \end{subfigure}
     ~
      \begin{subfigure}[b]{0.48\linewidth}
         \centering
        \includegraphics[width=\linewidth]{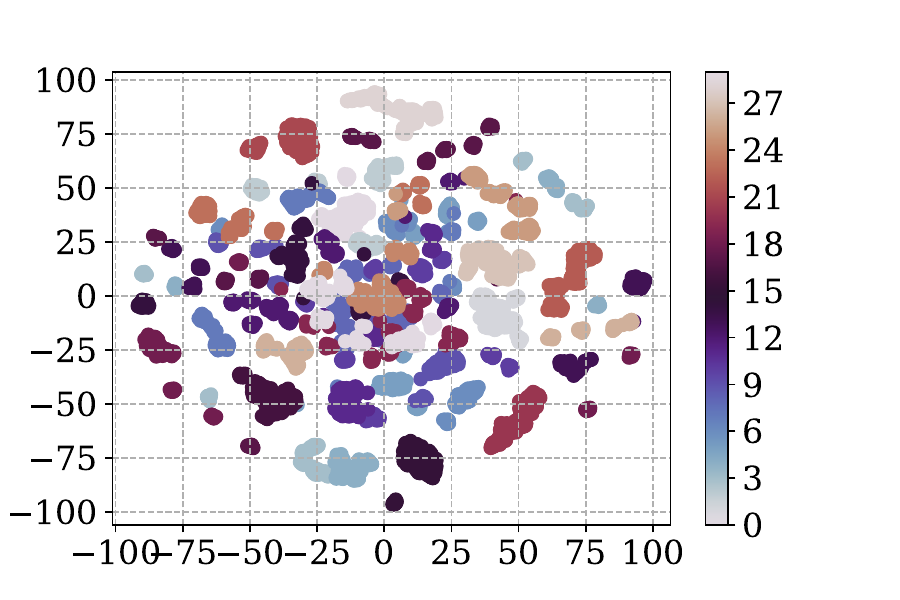}
        \caption{SHREC11.}
     \end{subfigure}
     \caption{t-SNE plots of test latent codes extracted from subdivision WrappingNet.}
     \label{fig:tsne_additional}
\end{figure}

We additionally show t-SNE plots for both Manifold10 and SHREC11 test sets of the learned latent codes from WrappingNet, in Fig.~\ref{fig:tsne_additional}, similar to the experiment in Sec.~4.2. Similar to the t-SNE plot of Manifold40, we see that the latent codes indeed cluster by class, demonstrating that WrappingNet learned global shape information in the latent codes. All figures (both t-SNE figures here and the one in the main text) use a perplexity parameter of 30. 

\subsubsection{Importance of Connectivity in the Sparse Regime}
As shown in the Sec.~4.2 of the main text, WrappingNet achieves similar classification performance as FoldingNet and TearingNet for the full-resolution subdivision meshes as well as the original manifold meshes. At the full resolution, the subdivision-remeshed Manifold40 has, on average, 3000 points per mesh; the original manifold meshes have a density of approximately 250 points per mesh. At these resolutions, the density of points on the surface of each shape is high enough such that the high-level information of the shape (\eg, its class membership) is preserved whether or not mesh connectivity is included. However, at much lower resolutions, a lack of mesh connectivity can significantly degrade the topology information of the shape. For example, a table top, which is flat, only needs 4 points (at the corners) with mesh connectivity to represent the flat surface; if connectivity is not available, 4 points at the corners will fail to represent the topology accurately. Since WrappingNet employs a mesh surface-based feature extraction, we expect it to maintain good performance even at lower point densities, where point cloud systems may begin to worsen in performance. 

We experimentally test this by training the models at a much lower point density of 50 points, on average. This corresponds to the point density of the simplified base mesh extracted during the subdivision remeshing. We use vanilla WrappingNet since there is no subdivision structure for the base mesh. For the point cloud autoencoders, we only use the vertex positions of the lower resolution meshes, and their decoder deforms a grid of similar size. Shown in Tab.~\ref{tab:density}, all models are able to maintain similar classification performance at the 250 point level (original manifold meshes) and the 300 point level (subdivision meshes). However, at 50 points, the performance of FoldingNet and TearingNet drops a significant amount by $\sim3\%$ for Manifold40 and $\sim4\%$ for Manifold10 when compared to the highest resolution. In comparison, WrappingNet only drops by $\sim1\%$ for Manifold40 and $\sim2\%$ for Manifold10. Overall, at higher resolutions, WrappingNet achieves comparable classification performance to the point cloud autoencoders, but at 50 points, WrappingNet achieves a \textit{clearly better} classification performance, demonstrating the importance of utilizing mesh connectivity at lower resolutions. 

\begin{table}[t]
  \centering\tiny
  \caption{Average classification performance from test latent representations using SVM with 5-fold cross-validation.}
    \begin{tabular}{c||c|c|c||c|c|c}
    \hline
    Dataset & \multicolumn{3}{c||}{Manifold10} & \multicolumn{3}{c}{Manifold40} \bigstrut\\
    \hline
    No. of Points & 50 & 250 & 3000 & 50 & 250 & 3000 \bigstrut\\
    \hline
    \hline
    FoldingNet & 86.8\% & 89.8\% & 91.0\% & 79.4\% & 82.5\% & 82.5\% \bigstrut[t]\\
    TearingNet & 86.7\% & 90.2\% & 90.8\% & 79.4\% & 82.9\% & 82.9\% \\
    WrappingNet & 89.0\% & 90.8\% & 90.9\% & 82.2\% & 83.0\% & 83.3\% \bigstrut[b]\\
    \hline
    \end{tabular}%
  \label{tab:density}%
\end{table}%

\subsection{Reconstructed Shape Analysis}

We describe in detail some of the metrics used to evaluate the reconstruction performance, and provide additional reconstruction examples following the latent interpolation and varying base graph experiment in Sec.~4.3. Then, we display a gallery of reconstructed shapes generated from WrappingNet.

\subsubsection{Reconstruction Metrics}
To evaluate reconstruction performance with point cloud autoencoders, we use Chamfer distance (CD), normals error (NE), and curvature preservation (CP). The latter two are inspired from surface-aware losses used in \cite{potamias2022}. All metrics are used to compare two sets of points $S_1, S_2$. In our setting, one represents the ground-truth vertex positions, and the other represents the reconstructed vertex positions. As mentioned in the main text, these metrics are all point-based (i.e., they do not consider connectivity information) for the purposes of comparing to the point cloud autoencoders. CD is computed as 
\begin{align}
    \mathsf{CD}(S_1, S_2) &:= \frac{1}{|S_1|} \sum_{\xv \in S_1} \min_{\xv' \in S_2} \|\xv - \xv'\|_2^2 \nonumber \\ &+ \frac{1}{|S_2|} \sum_{\xv \in S_2} \min_{\xv' \in S_1} \|\xv - \xv'\|_2^2. 
\end{align}
NE metric is computed as
\begin{align}
    \mathsf{NE}(S_1, S_2) &:= \frac{1}{|S_1|} \sum_{\substack{\xv \in S_1 \\ \xv'=\underset{\yv \in S_2}{\argmin} \|\xv-\yv\|_2^2}} 1 - \frac{\nv_{\xv} \cdot \nv_{\xv'}}{\|\nv_\xv\|_2 \|\nv_{\xv'}\|_2} \nonumber \\ &+ \frac{1}{|S_2|} \sum_{\substack{\xv \in S_2 \\ \xv'=\underset{\yv \in S_1}{\argmin} \|\xv-\yv\|_2^2}} 1 - \frac{\nv_\xv \cdot \nv_{\xv'}}{\|\nv_\xv\|_2 \|\nv_{\xv'}\|_2}, 
\end{align}
where $\nv_\xv$ is the normal vector at point $\xv$. Normals are estimated via covariance methods if they do not exist. CP uses the covariance-based curvature of a $k$-nearest neighborhood around each point defined in \cite[Sec.~3.1]{potamias2022eccv}. In our experiments, we set $k=15$ for calculating the CP metric.

\subsubsection{Additional Qualitative Reconstruction Comparison}
We show additional qualitative topology comparisons with TearingNet reconstructed point clouds and meshes in Tab.~\ref{tab:topology_comparison_supp}. Again, as in the main text, WrappingNet clearly does a better job in preserving the mesh topology, which supports the lower reconstruction loss for the surface-aware metrics shown in Tab.~3 of the main text.

\begin{table}[t]
  \centering\scriptsize
  \caption{Topology comparison of reconstructed shapes. Both a point cloud and mesh rendering are provided for each object.}
    \begin{tabular}{c|c||ccc}
    \hline
          &       & Ground Truth    & TearingNet & WrappingNet \bigstrut\\
    \hline
    \multirow{2}[2]{*}{\begin{sideways}Monitor\end{sideways}} & \begin{sideways}\phantom{iiiiiii}Mesh\end{sideways} & 
    \includegraphics[width=0.18\linewidth]{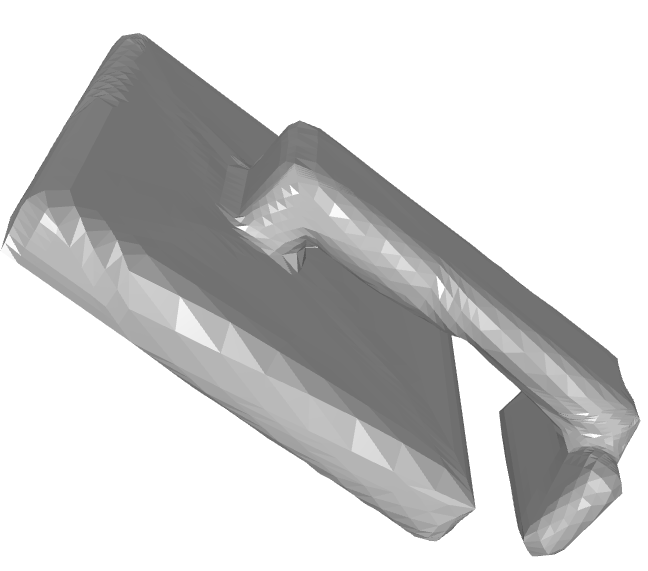} & \includegraphics[width=0.17\linewidth]{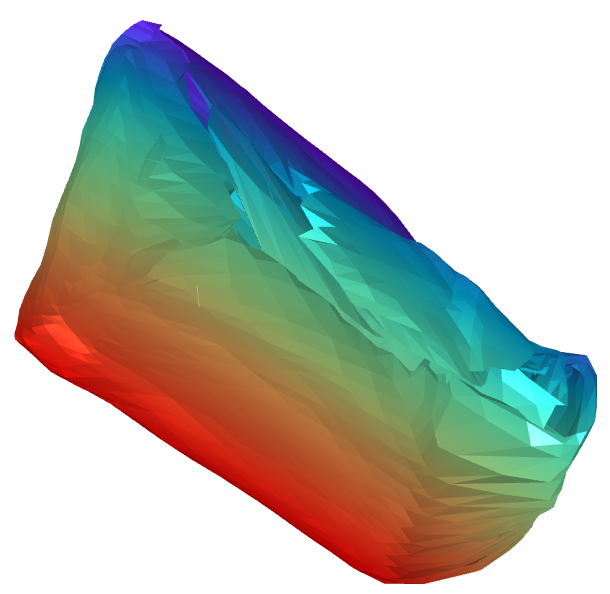} & \includegraphics[width=0.18\linewidth]{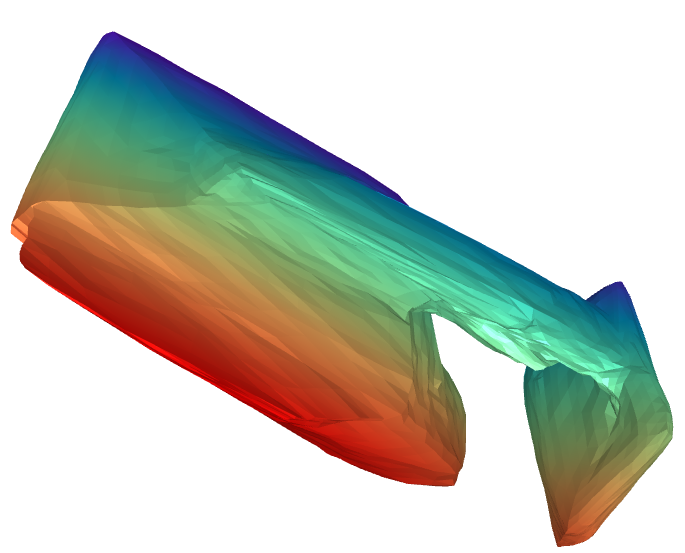} \bigstrut[t] \vspace{-.5em} \\
          & \begin{sideways}Point Cloud\end{sideways} & \includegraphics[width=0.18\linewidth]{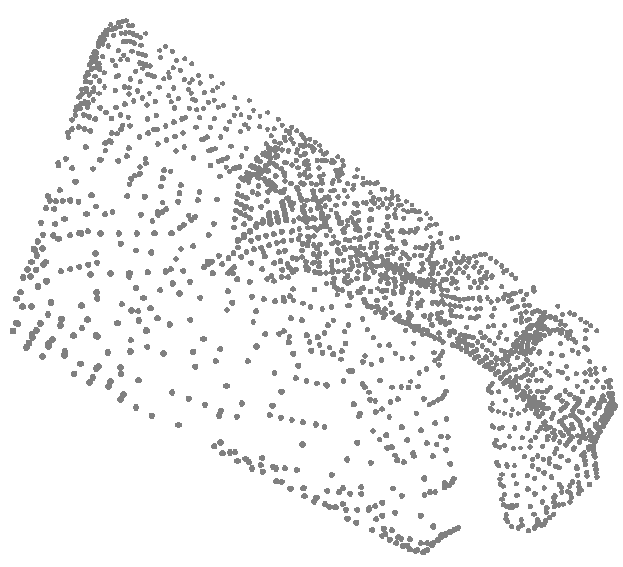} & \includegraphics[width=0.18\linewidth]{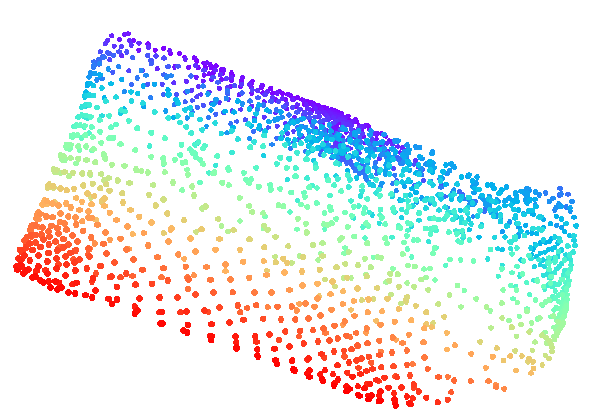} & \includegraphics[width=0.18\linewidth]{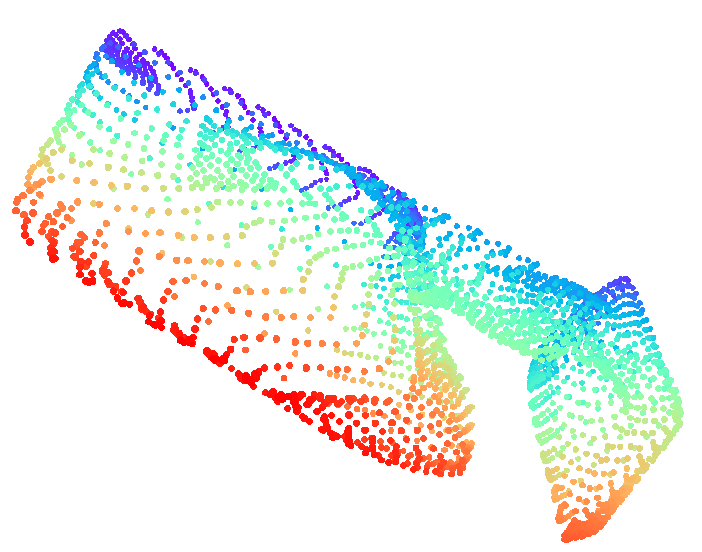} \bigstrut[b]\\
    \hline
    \hline
    \multirow{2}[2]{*}{\begin{sideways}Table\end{sideways}} & \begin{sideways}\phantom{iiiiii}Mesh\phantom{iii}\end{sideways} & \includegraphics[width=0.15\linewidth]{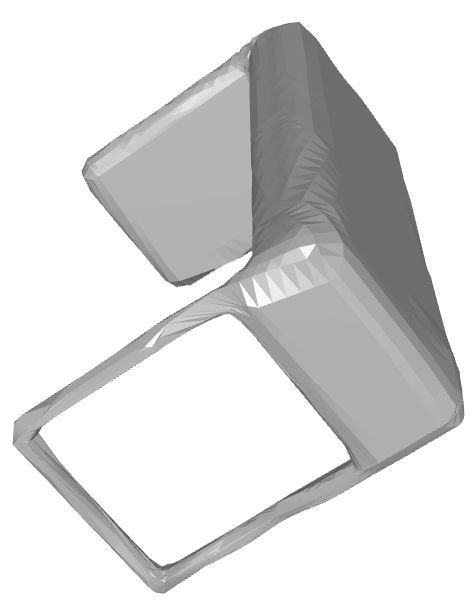} & \includegraphics[width=0.14\linewidth]{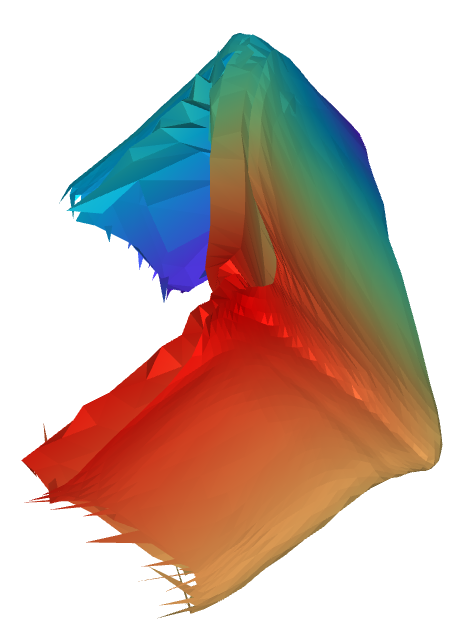} & \includegraphics[width=0.15\linewidth]{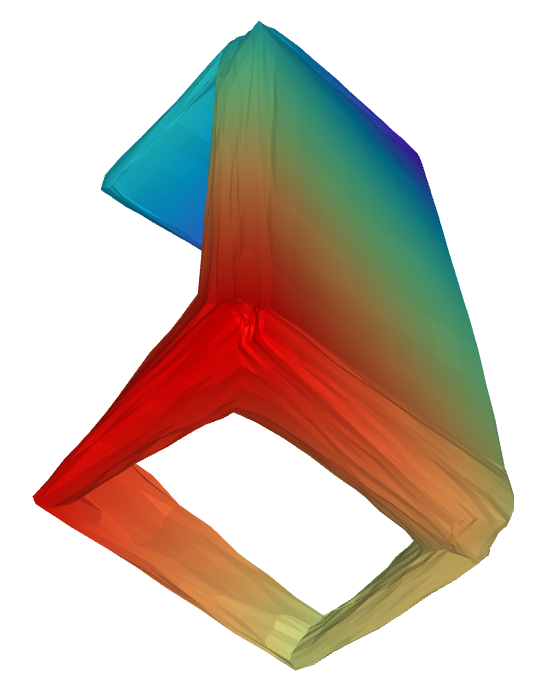} \bigstrut[t]  \vspace{-.4em}  \\
          & \begin{sideways}\phantom{iii}Point Cloud\end{sideways} & \includegraphics[width=0.15\linewidth]{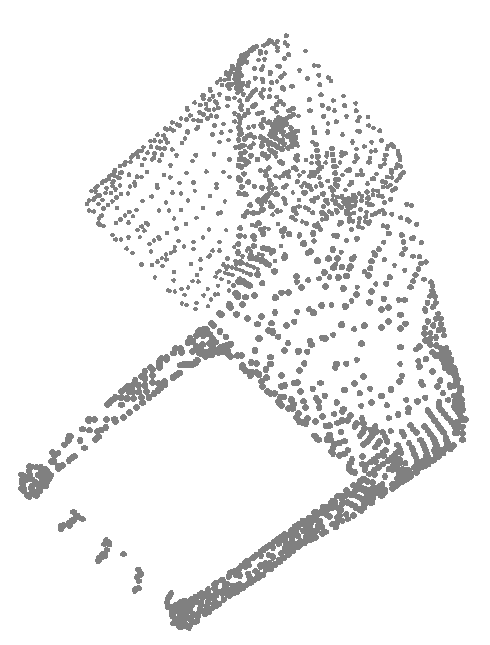} & \includegraphics[width=0.15\linewidth]{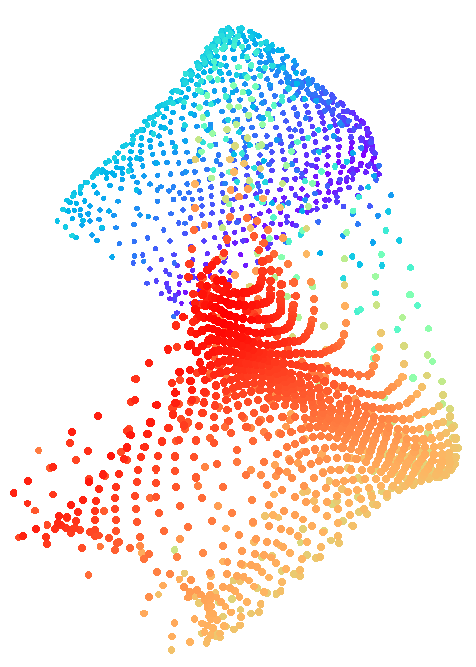} & \includegraphics[width=0.15\linewidth]{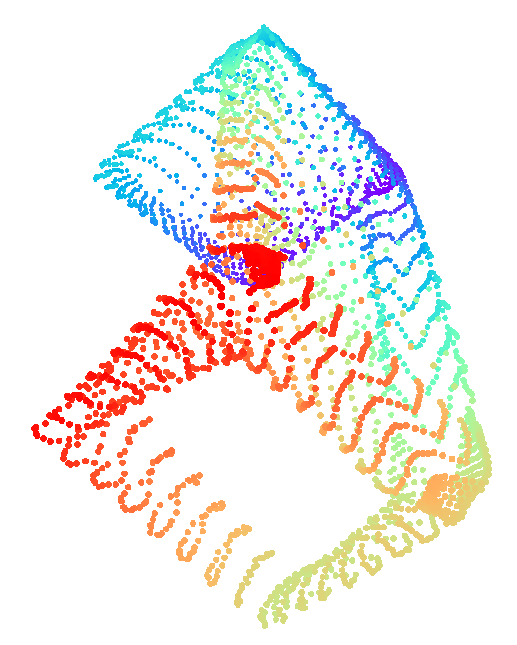} \bigstrut[b]\\
    \hline
    \hline
    \multirow{2}[2]{*}{\begin{sideways}Bathtub\end{sideways}} & \begin{sideways}\phantom{iiiiiiii}Mesh\phantom{iii}\end{sideways} & \includegraphics[width=0.15\linewidth]{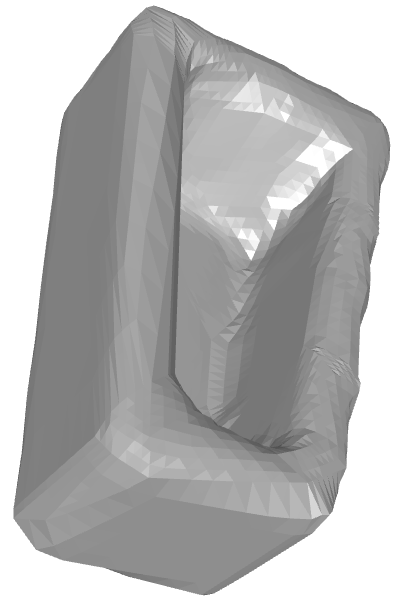} & \includegraphics[width=0.15\linewidth]{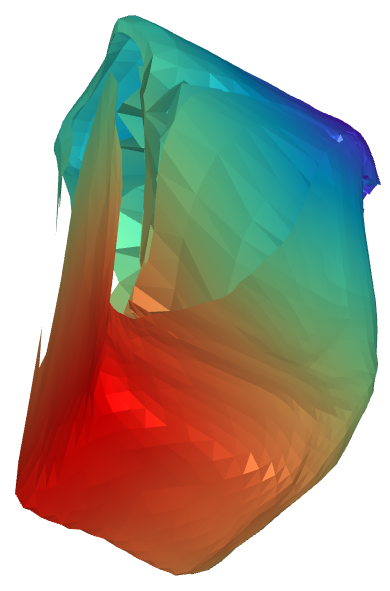} & \includegraphics[width=0.15\linewidth]{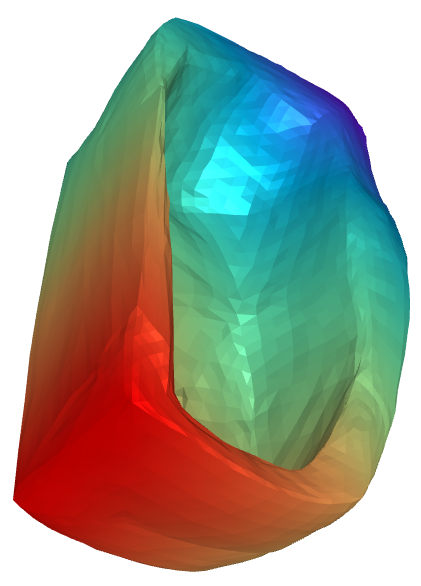} \bigstrut[t]  \vspace{-.4em}  \\
          & \begin{sideways}\phantom{iiii}Point Cloud\end{sideways} & \includegraphics[width=0.15\linewidth]{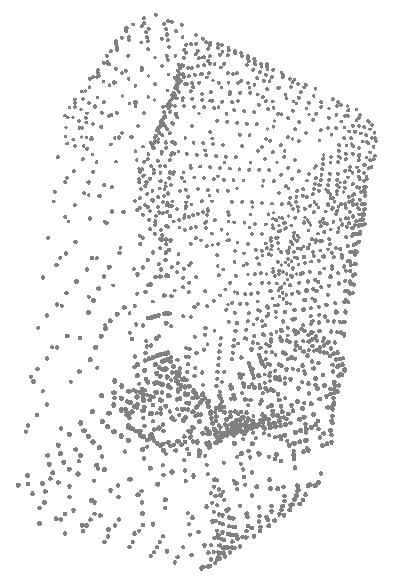} & \includegraphics[width=0.15\linewidth]{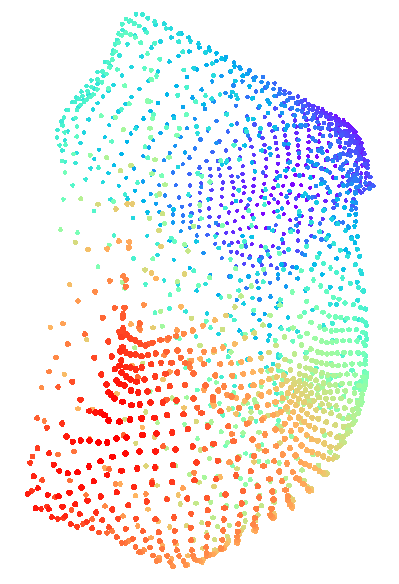} & \includegraphics[width=0.15\linewidth]{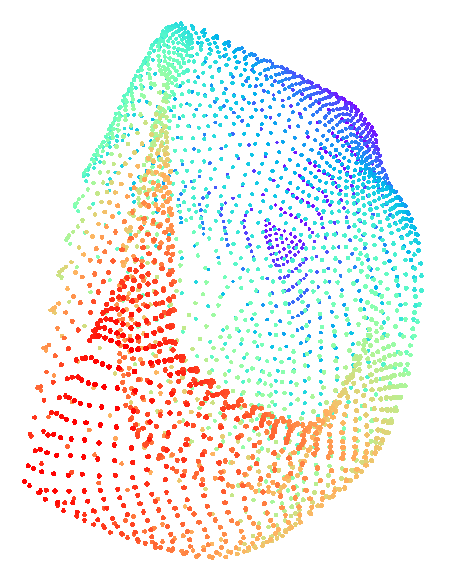} \bigstrut[b]\\
    \hline
    \end{tabular}%
  \label{tab:topology_comparison_supp}%
\end{table}%

\subsubsection{Latent Interpolation}
\begin{figure}[h]
    \centering
     \begin{subfigure}[b]{0.5\textwidth}
         \centering
        \includegraphics[width=0.85\linewidth]{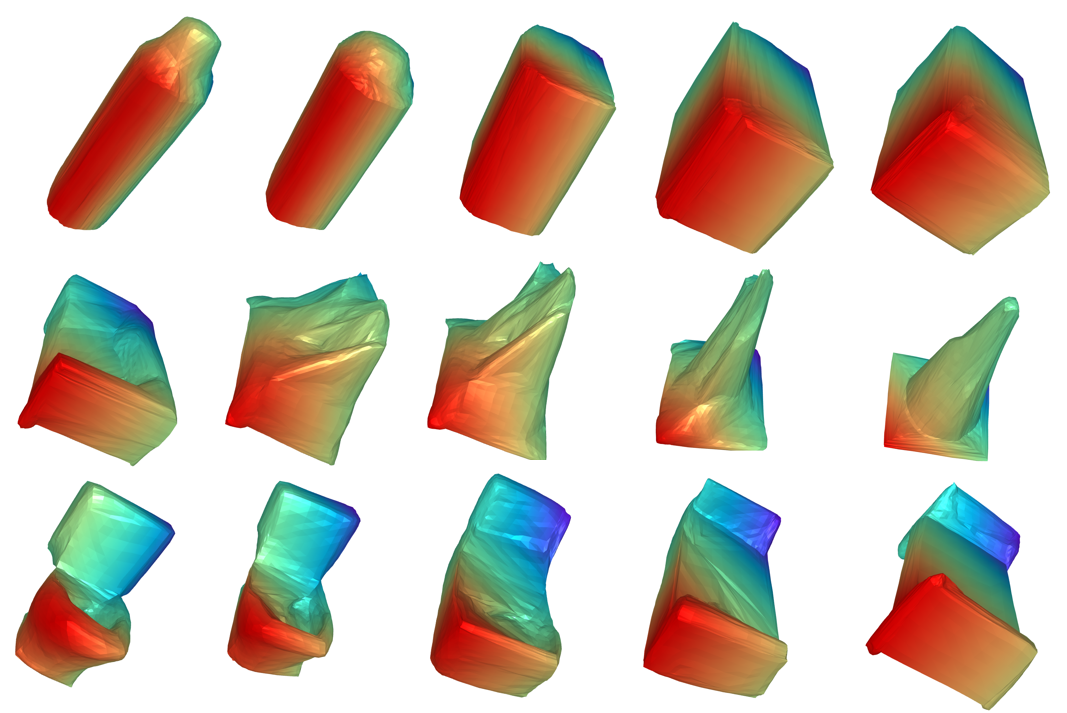} \vspace{-.3em}
        \caption{\textit{Inter}-class. $1^{\text{st}}$ row: bottle $\rightarrow$ glass box. $2^{\text{nd}}$ row: sofa $\rightarrow$ cone. $3^{\text{rd}}$ row: toilet $\rightarrow$ bed.}
     \end{subfigure}
      \begin{subfigure}[b]{0.5\textwidth}
         \centering
        \includegraphics[width=0.85\linewidth]{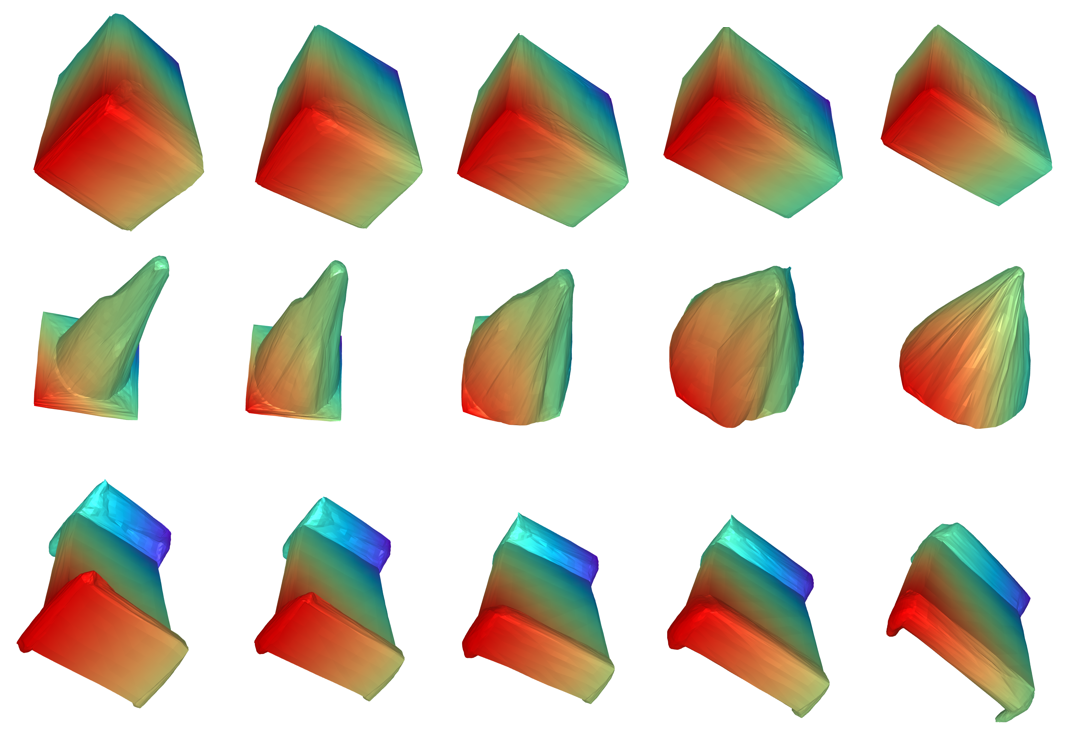} \vspace{-.3em}
        \caption{\textit{Intra}-class. $1^{\text{st}}$ row: glass box. $2^{\text{nd}}$ row: cone.
        $3^{\text{rd}}$ row: bed.}
     \end{subfigure}
     \caption{Latent space interpolation. Base graph in use is generated from WrappingNet encoder on meshes from the left side. }
     \label{fig:latent_interp_supp}
\end{figure}

The additional reconstructed shapes generated via interpolating the latent codes can be seen in Fig.~\ref{fig:latent_interp_supp}, which follows the procedure described in Sec.~4.3.

\subsubsection{Reconstruction from Varying Base Graphs}
The additional results demonstrating the reconstruction of shapes from varying the base graphs can be found in Tab.~\ref{tab:mix_match_supp}. Again, we see that the codeword maintains global shape information, whereas the base graph influences the topology of the reconstructed shape. When the genus is mis-matched between the codeword and the base graph (gray-scale reconstructions), the decoder tries its best to fit the mis-matched topology to the global shape embedded in the codeword. 

{\emph{In particular}, using the codeword from a higher genus object and the base graph from a lower genus, the reconstructed meshes still appear pleasant (the two gray meshes in the 3rd row in Tab.~\ref{tab:mix_match_supp}). On the contrary, if the codeword is from a lower genus object and the base graph is from a higher genus, the reconstruction is much less meaningful (the two gray meshes in the 3rd column in Tab.~\ref{tab:mix_match_supp}). This provides additional evidence that the codeword is more dominant (than the base graph) in rebuilding the global shape, that is aligned with the intention of our autoencoder design.}

\begin{table}[t]
  \centering\scriptsize
  \caption{Reconstructions from various base graphs \& latent codes. }
    \begin{tabular}{rc|ccc}
    \hline
          &       & \multicolumn{3}{c}{Base graph comes from} \vspace{-.3em} \bigstrut[t]\\
          &       \phantom{\includegraphics[width=0.155\linewidth]{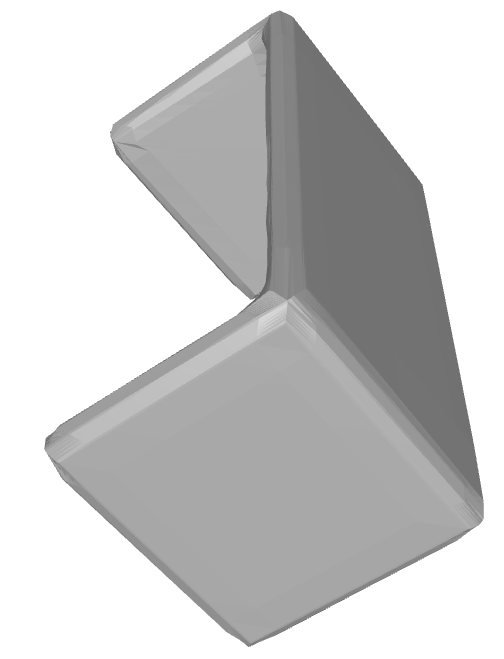}}& \includegraphics[width=0.15\linewidth]{diagrams/figures_decoder_supp2/c1_gt.png}    & \includegraphics[width=0.14\linewidth]{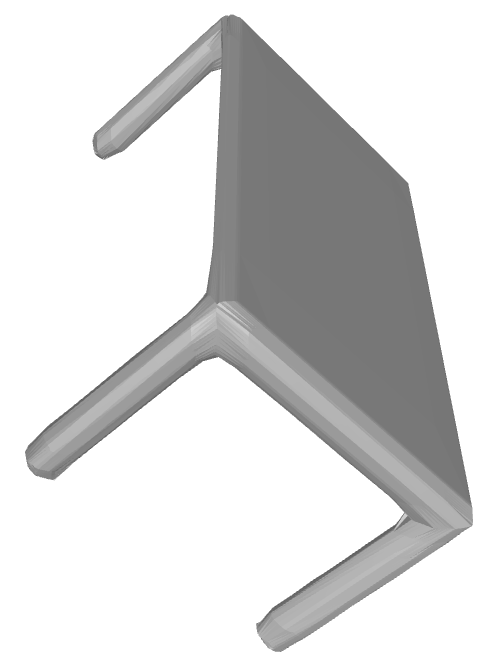}    & \includegraphics[width=0.15\linewidth]{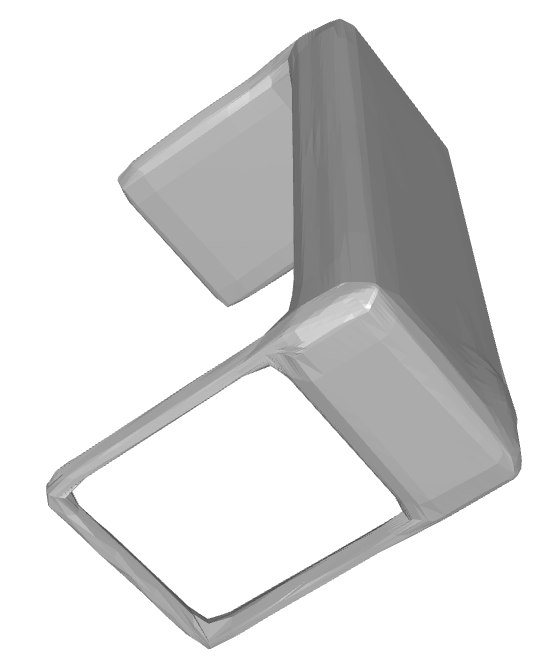} \vspace{-.3em} \bigstrut[b]\\
    \hline
    \multirow{3}[2]{*}{\begin{sideways}\phantom{aaaa}Latent codes comes from\end{sideways}} & \includegraphics[width=0.15\linewidth]{diagrams/figures_decoder_supp2/c1_gt.png}   & \includegraphics[width=0.15\linewidth]{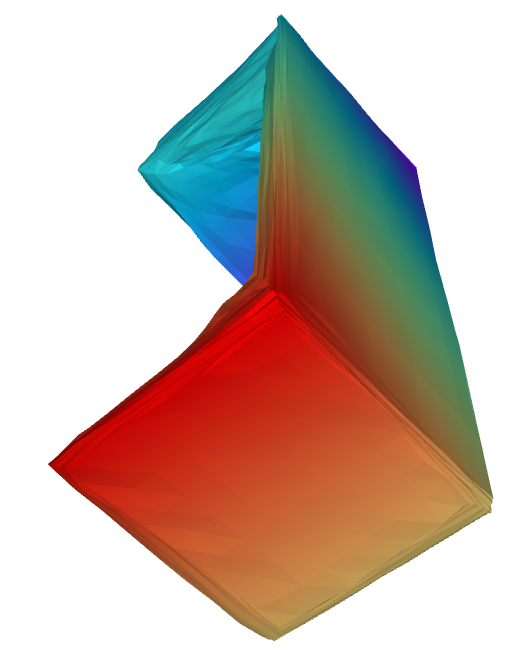}  & \includegraphics[width=0.12\linewidth]{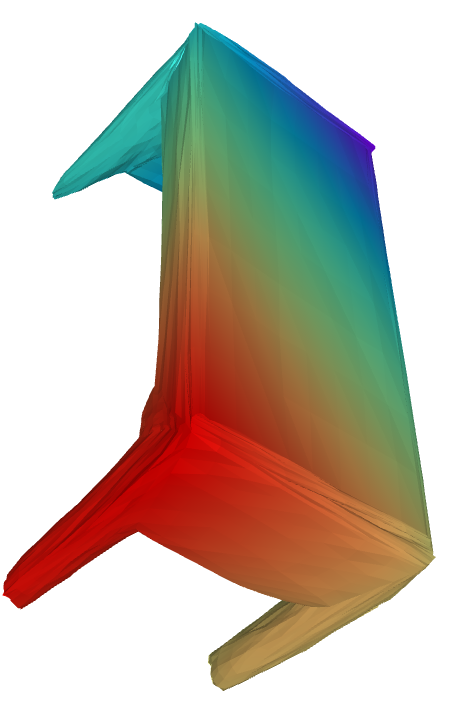}  & \includegraphics[width=0.13\linewidth]{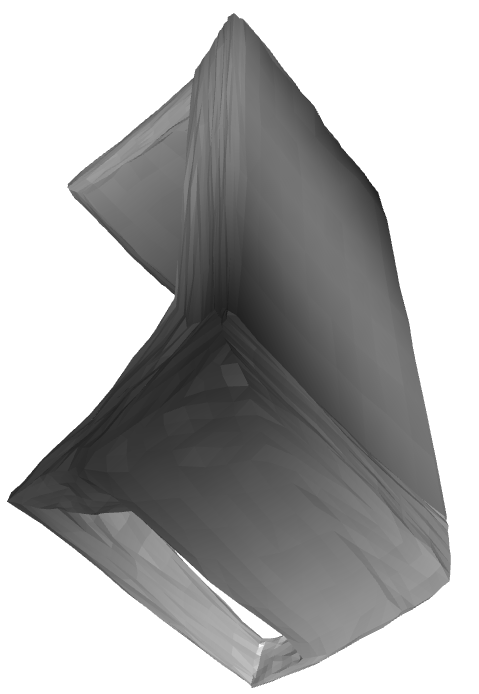} \phantom{\includegraphics[width=0.001\linewidth,height=0.17\linewidth]{diagrams/topology_comparison/2_gt_pc.png}} \bigstrut[t]\\
          & \includegraphics[width=0.15\linewidth]{diagrams/figures_decoder_supp2/c2_gt.png}   & \includegraphics[width=0.15\linewidth]{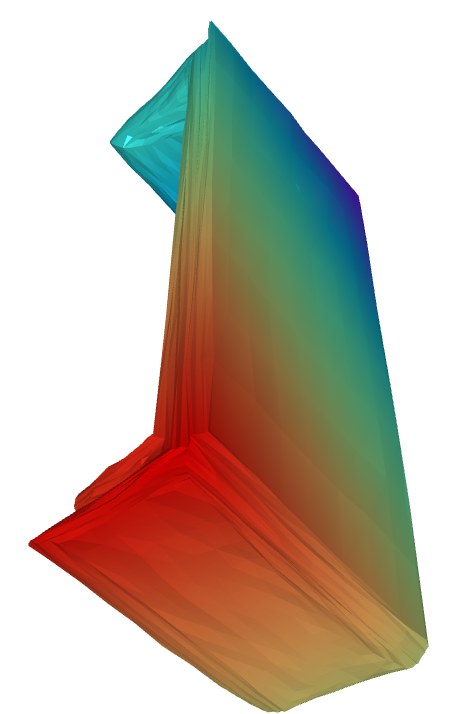}  & \includegraphics[width=0.15\linewidth]{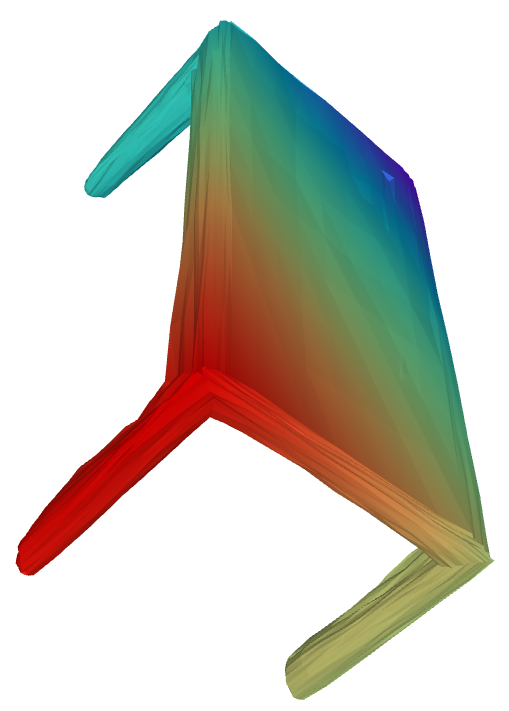}  & \includegraphics[width=0.14\linewidth]{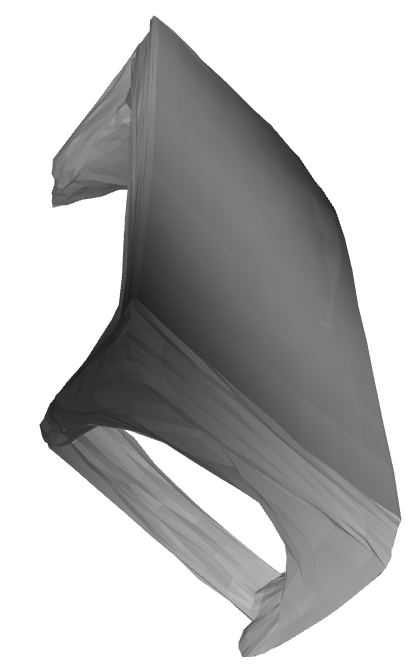} \\
          & \includegraphics[width=0.15\linewidth]{diagrams/figures_decoder_supp2/c3_gt.png}   & \includegraphics[width=0.15\linewidth]{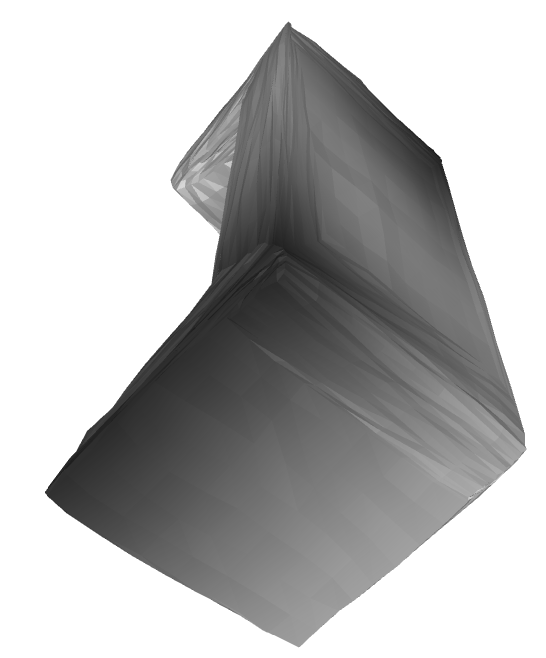}  & \includegraphics[width=0.15\linewidth]{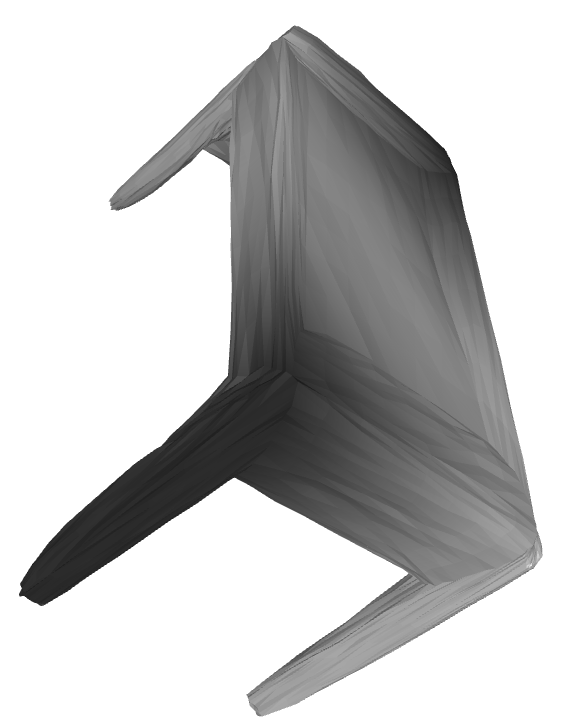}  & \includegraphics[width=0.15\linewidth]{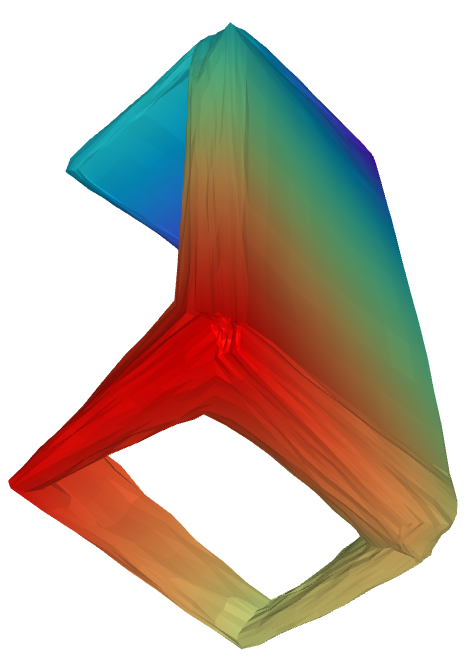} \bigstrut[b]\\
    \hline
    \end{tabular}%
  \label{tab:mix_match_supp}%
  \vspace{-10pt}
\end{table}%

\subsubsection{Gallery of reconstructions}
\begin{figure*}[t]
    \centering
    \includegraphics[width=1.0\linewidth]{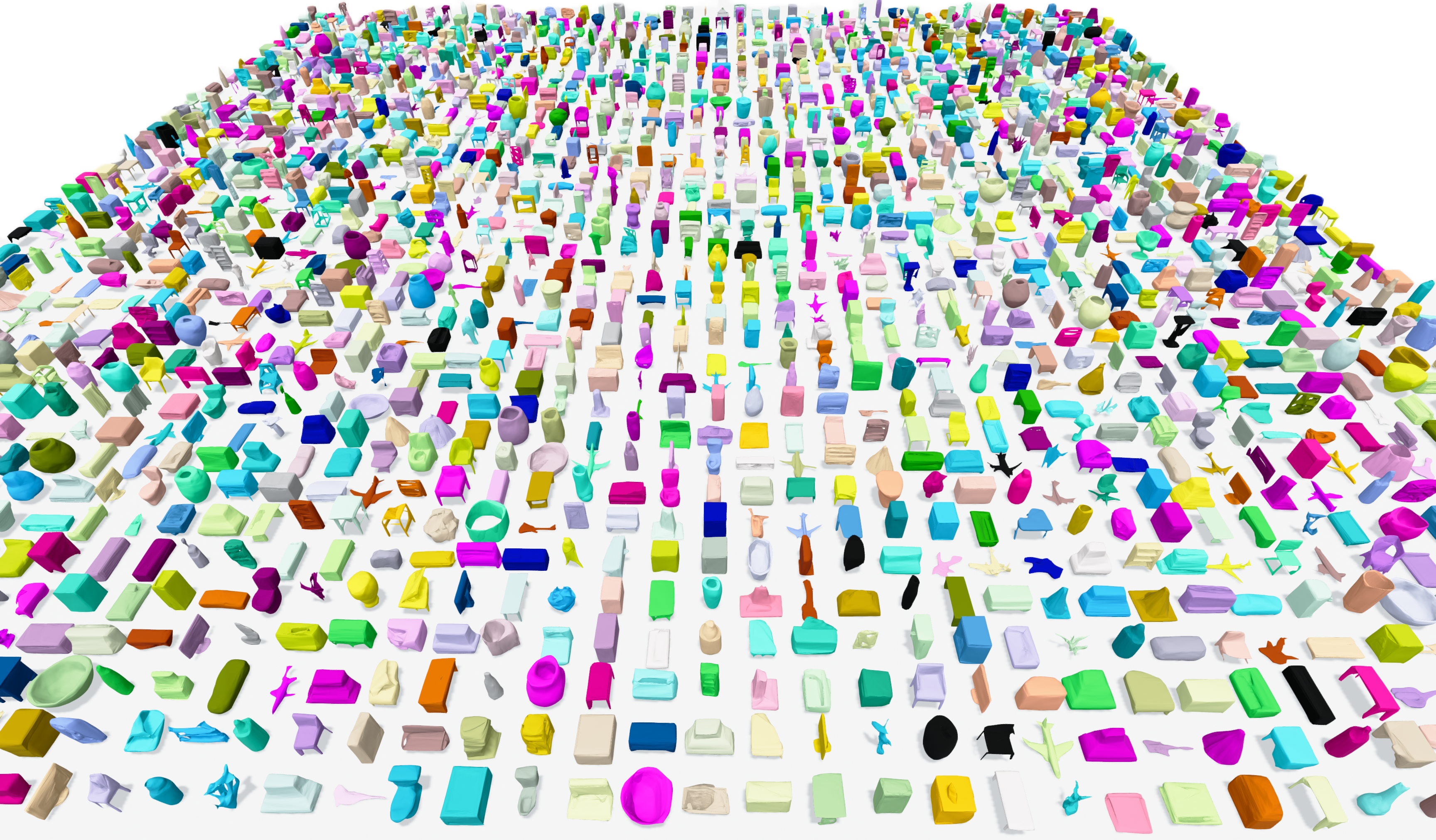}
    \caption{Gallery of reconstructed meshes. WrappingNet extracts a \textit{shared latent space} across all such heterogeneous meshes.}
    \label{fig:gallery}
\end{figure*}

In Fig.~\ref{fig:gallery}, we display a gallery of reconstructed meshes from WrappingNet, on the Manifold40 test set. WrappingNet is able to learn a \textit{shared latent space across all such meshes}, despite the heterogeneity of the dataset. 

\subsection{Ablation Study}
\begin{figure}[t]
    \centering
     \begin{subfigure}[b]{0.48\linewidth}
         \centering
        \includegraphics[width=\linewidth]{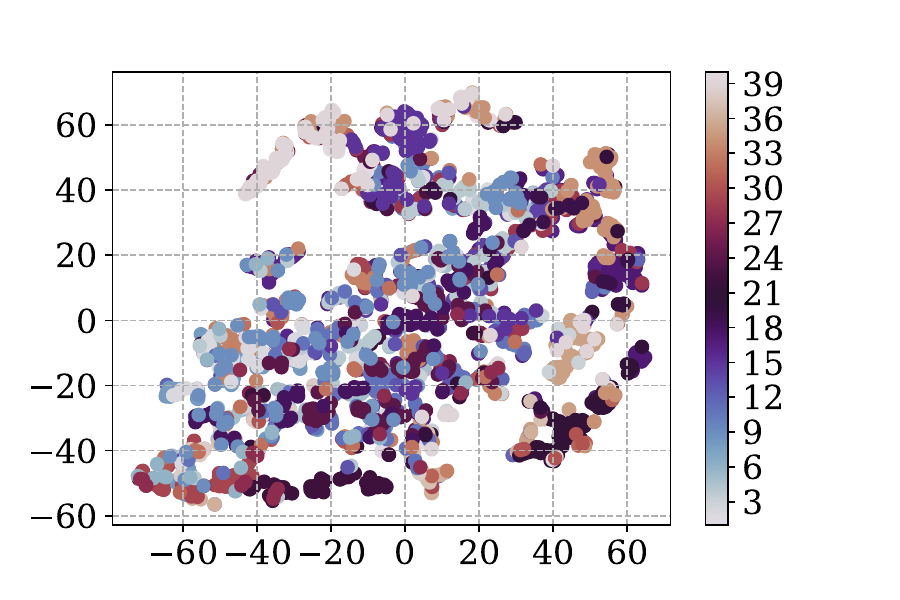}
        \caption{Send base mesh: no wrapping and no sphere matching.}
     \end{subfigure}
     \hfill
      \begin{subfigure}[b]{0.48\linewidth}
         \centering
        \includegraphics[width=\linewidth]{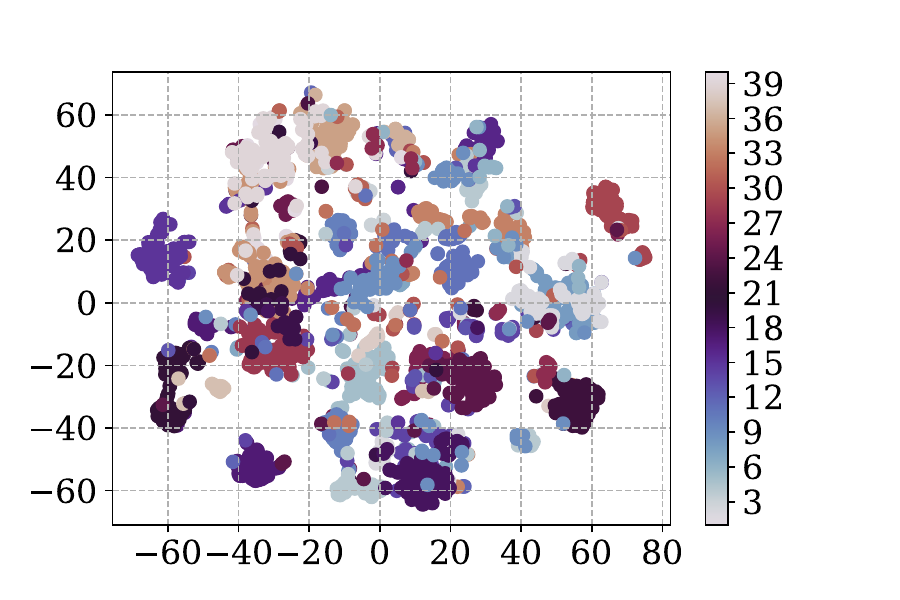}
        \caption{Send $\MAS$: with wrapping but without sphere matching.}
     \end{subfigure}
     \caption{t-SNE plots of latent codes extracted from Manifold40 on different configurations (ablation study) of WrappingNet. Configuration with both wrapping/unwrapping and matching can be found in Fig.~5 of the main text. }
     \label{fig:tsne_ablation}
\end{figure}

We demonstrate t-SNE figures of extracted latent codes of the first two configurations in the ablation study (Sec.~4.4), shown in Fig.~\ref{fig:tsne_ablation}. These figures further support the benefit of wrapping operations in pushing the global shape information into the latent code. As shown, the first configuration (sending the base mesh) fails to capture any class-related clustering phenomenon in the latent codes, since the latent code and base mesh are entangled. In this setting, the latent code is merely used to upsample the base mesh. For the second configuration (sending $\MAS$), where we use wrapping operations but no matching, there is more disentanglement. The full WrappingNet model with both wrapping operations and matching (Fig.~5 of main text) displays the most disentanglement achieved.  

\section{WrappingNet Implementation Details}
The code will be made public upon full publication. 
\subsection{Sphere Matching Details}
Recall that the sphere grid of points on the sphere $\{(\bar{x}_i, \bar{y}_i, \bar{z}_i)\}_{i=1}^N$ is defined by the Fibonacci lattice \cite{fibonaccilattice}.
Let $\bar{\sv}_i = (\bar{x}_i, \bar{y}_i, \bar{z}_i)$ be the $i$-th point on the sphere grid, and let $\sv_i = (x_i, y_i, z_i)$ be the $i$-th point of the output of the UnWrapping layers. Since UnWrapping has been pretrained to approximately output unit spheres, we expect the $\vect{s}_i$'s to approximately lie on the surface of a sphere. However, if we compute nearest neighbors between $\bar{\sv}_i$'s and $\sv_i$'s, they may not be in correspondence with each other. To resolve this, we use the assignment problem to compute a correspondence between the two sets of points. 

Let $\Cv_{i,j} \triangleq \|\sv_i - \bar{\sv}_j\|_2^2$ be the pairwise squared Euclidean distance between the $i$-th point in the UnWrapping output and the $j$-th point in the sphere grid. Since the number of points in the UnWrapping output and the sphere grid do not match, we solve the unbalanced variant of the assignment problem \cite{peyre2019computational}
\begin{equation}
    \hat{\sigma} = \argmin_{\sigma : \{1,\dots,N'\} \rightarrow \{1,\dots,N\}} \sum_{i=1}^{N'} \Cv_{i, \sigma(i)}, 
    \label{eq:monge}
\end{equation}
where $N'$ is the number of points the UnWrapping output. This can be solved using the Hungarian algorithm \cite{Kuhn1955Hungarian}, which will return a bijection $\hat{\sigma}$.

Once the UnWrapping output gets matched with the sphere grid, the base graph (base mesh connectivity) gets re-indexed to be in the ordering of the sphere grid via $\hat{\sigma}$, and sent to the decoder. To enforce the vertex-to-vertex loss between the reconstructed mesh and ground-truth mesh, the matching $\hat{\sigma}$ can be used to ensure the input mesh and reconstructed mesh are in the same node ordering (either that of the original mesh or of the sphere grid). We clarify that the use of the matching $\hat{\sigma}$ to align the input mesh and reconstructed mesh is only needed to enforce the loss during training; during inference, all that is needed is to use the matching to re-index the base graph before sending it to the decoder. 

For subdivision meshes, the sphere matching is performed at the lowest level of subdivision (the base mesh). Since subdivision meshes' face and vertex indexing is deterministic and completely determined by the base mesh's indexing, the bijection $\hat{\sigma}$ at the base mesh level is sufficient to compute the corresponding bijection at any level of subdivision.

\subsection{Face Feature Initialization}
\label{sec:init_features}
We wish to ensure that the input features are invariant to the ordering of nodes and faces, and the global position or orientation of the face. Hence, we choose the input face features to be the normal vector of the face, the face area, and a vector containing curvature information of the face, which is defined as follows. For face $i$, let $j_0, j_1, j_2$ denote the face indices of its 3 neighbors. The curvature vector is simply $\bv^{c}_i - \frac{1}{3}(\bv^{c}_{j_0}+\bv^{c}_{j_1}+\bv^{c}_{j_2})$ where $\bv^{c}_i, \bv^{c}_{j_0}, \bv^{c}_{j_1}, \bv^{c}_{j_2}$ are the centroids of the faces respectively. Thus, we have a total of 7 input features. These are used for the meshes that are inputs to the Wrapping and UnWrapping modules. For the Wrapping modules at the decoder, they are concatenated with the copied codeword. 

\subsection{Model Architecture}

\begin{figure*}[t]
    \centering
    \includegraphics[width=1\linewidth]{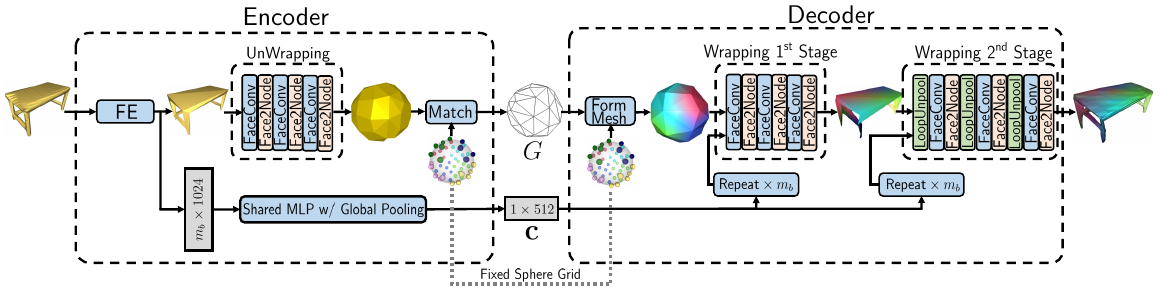}
    \caption{Subdivision-enhanced WrappingNet architecture. The input mesh is assumed to be a subdivision mesh with base mesh containing $m_b$ faces. Loop pooling and subdivision are incorporated into the feature extractor (FE) at the encoder, and 2nd Wrapping stage at the decoder. }
    \label{fig:E2E_subdiv}
\end{figure*}

All FaceConv layers use a kernel size of 3 and dilation of 1 defined in \cite{subdivnet}. The convolution operation, defined there as well, is order-invariant, since it uses 4 learnable weights $w_0,\dots,w_3$ to apply the convolution, which is given by
\begin{align}
    \vect{F}'_i = w_0 \vect{F}_i + w_1 \sum_{j \in \mathcal{N}_i}  \vect{F}_j &+ w_2 \sum_{j \in \mathcal{N}_i} | \vect{F}_{j+1}- \vect{F}_j| \nonumber \\ &+ w_3 \sum_{j \in \mathcal{N}_i} | \vect{F}_i -  \vect{F}_j|,
\end{align}
where $\vect{F}_i$ is the face feature on face $i$ and $\mathcal{N}_i$ is the set of three face indices adjacent to face $i$ corresponding to its edges. This is done for each input-output channel pair.

We use the PyTorch framework \cite{pytorch} for the implementation of WrappingNet, along with PyTorch Geometric \cite{pyg} to implement sparse batching of meshes. Multi-layer perceptron (MLP) layers are applied face-wise. All MLP and FaceConv layers use biases. ReLU activations exist after each fully connected layer in MLPs except for the last layer. MLPs in the Face2Node modules have 2 layers, and follow the same style as other MLPs. At the encoder, the feature extractor uses 4 FaceConv layers with hidden dimension of 128, and the shared MLP has 4 hidden layers of dimension 1024. The Wrapping module uses a hidden dimension of 64. At the decoder, the UnWrapping module uses a hidden dimension of 128. 

\subsection{Subdivision Enhanced WrappingNet}

In the subdivision-enhanced version of WrappingNet, the primary difference is that Loop pooling and subdivision is applied at the encoder and decoder in order to incorporate hierarchical feature extraction, as well as perform the matching on a smaller base mesh. The feature extractor (FE) now applies Loop pooling after each of its face convolutions layers, and outputs a feature map on the base mesh, which contains $m_b$ faces. UnWrapping and codeword pooling now takes place on the base mesh. At the decoder, we now have two Wrapping modules. The first stage is the same as before, with no Loop subdivision; this recovers an approximate base mesh. Then, a second Wrapping stage is applied, which interleaves Loop subdivision into the layers\footnote{The subdivided points' vertex positions take the midpoints of the edges they lie on.}. In all experiments, we use 3 levels of subdivision, which is reflected in our architecture; the FE uses 3 Loop pooling layers, and Wrapping 2nd stage uses 3 Loop subdivision (unpooling) layers.

\subsection{Model Size and Efficiency}

A forward pass during evaluation (without accumulating gradients), consisting of encoding and decoding, using WrappingNet takes $\approx 254$ ms per mesh on the Manifold40 test set. This is evaluated on an NVIDIA GeForce RTX 2080 Ti GPU. The total number of parameters is $\approx 4.6$M, which is on the same order of magnitude with \cite{pang2021tearingnet, foldingnet}.

\subsection{Remeshing Details for Generating Subdivision Meshes}
In this section, we elaborate on the remeshing methods used for SHREC11 and Manifold40.  All remeshing techniques that generate subdivision meshes have the same overall process: 
\begin{enumerate}[(1)]
    \item Apply a series of edge collapses to the original mesh in order to generate a low-resolution base mesh.
    \item Subdivide the base mesh $L$ times, with new vertex positions on the midpoints of edges.
    \item Project all vertex positions onto the original mesh to approximately recover the shape at subdivision level $L$.
\end{enumerate}
The primary differences among remeshing methods are in steps (1) and (3). For SHREC11, we use the subdivision meshes generated from \cite{subdivnet}, who use the MAPS \cite{maps} method. For Manifold40, we use the original Manifold40 meshes from \cite{subdivnet} (not subdivision), and remesh them using the improved version of MAPS from \cite{neuralsubdiv}. In this case, for step (1) we use quadric error simplification \cite{qem} to simplify down to a base mesh containing approximately 50 vertices, and for step (3) we use the parametrization between the subdivided mesh and the original mesh explained in \cite{neuralsubdiv}. 

\end{document}